\documentclass[lettersize,journal]{IEEEtran}

\ifCLASSOPTIONcompsoc
\usepackage[nocompress]{cite}
\else
\usepackage{cite}
\fi

\usepackage{amsmath,amsfonts}
\usepackage{algorithmic}
\usepackage{algorithm}
\usepackage{array}
\usepackage{textcomp}
\usepackage{stfloats}
\usepackage{url}
\usepackage{verbatim}
\usepackage{graphicx}
\usepackage{cite}
\usepackage{amsmath,amsfonts,bm}

\def\eqref#1{equation~\ref{#1}}

\def\1{\bm{1}}

\DeclareMathAlphabet{\mathsfit}{\encodingdefault}{\sfdefault}{m}{sl}
\SetMathAlphabet{\mathsfit}{bold}{\encodingdefault}{\sfdefault}{bx}{n}

\DeclareMathOperator*{\argmin}{arg\,min}

\usepackage[colorlinks,citecolor=blue,urlcolor=blue]{hyperref}
\usepackage{url}
\usepackage{color,soul}
\usepackage{multirow}
\usepackage{epsfig}
\usepackage{booktabs}
\usepackage{subcaption}
\usepackage{mathtools}

\usepackage[table]{xcolor}
\definecolor{highlightgreen}{HTML}{39b54a}

\usepackage{multirow}

\begin{document}

\title{Dynamic Context-oriented Decomposition for Task-aware Low-rank Adaptation with Less Forgetting and Faster Convergence} 

\author{Yibo Yang, 
	Sihao Liu, Chuan Rao, Bang An, Tiancheng Shen,\\
	Philip H.S. Torr, 
	Ming-Hsuan Yang, 
	and Bernard Ghanem
\thanks{Yibo Yang, Sihao Liu, Chuan Rao, Bang An, and Bernard Ghanem are with King Abdullah University of Science and Technology, Saudi Arabia.}
\thanks{Tiancheng Shen and Ming-Hsuan Yang are with University of California, Merced, USA.}   
\thanks{Philip H.S. Torr is with the University of Oxford, U.K.}  
\thanks{Corresponding authors: Yibo Yang (e-mail: yibo.yang93@gmail.com), Bernard Ghanem (e-mail: bernard.ghanem@kaust.edu.sa)}
}



\maketitle

\begin{abstract}
Conventional low-rank adaptation methods build adapters without considering data context of either the downstream task to learn or the pre-trained knowledge that needs preservation. 
As a result, the initialized adapters are task-agnostic, leading to sub-optimal fine-tuning performance and severe forgetting of inherent world knowledge. 
In this paper, we propose context-oriented decomposition adaptation (CorDA), a novel method that initializes adapters in a task-aware manner. 
Concretely, we develop context-oriented singular value decomposition, where we collect covariance matrices of input activations for each linear layer using sampled data from the target task, and apply SVD to the product of weight matrix and its corresponding covariance matrix. 
By doing so, the task-specific capability is compacted into the principal components. 
Thanks to the task awareness, our method enables two optional adaptation modes, knowledge-preserved mode (KPM) and instruction-previewed mode (IPM), providing flexibility to choose between freezing the principal components to preserve their associated knowledge or adapting them to better learn a new task. 
We further develop CorDA++ by deriving a metric that reflects the compactness of task-specific principal components, and then introducing dynamic covariance selection and dynamic rank allocation strategies based on the same metric. 
The two strategies provide each layer with the most representative covariance matrix and a proper rank allocation. 
Experimental results show that CorDA++ outperforms CorDA by a significant margin. 
CorDA++ in KPM not only achieves better fine-tuning performance than LoRA, but also mitigates the forgetting of pre-trained knowledge in both large language models and vision language models. 
For IPM, our method exhibits faster convergence, \emph{e.g.,} 4.5x speedup over QLoRA, and improves adaptation performance in various scenarios, outperforming strong baseline methods. 
Our method has been integrated into the PEFT library developed by Hugging Face at \url{https://github.com/huggingface/peft}, with available examples at \href{https://github.com/huggingface/peft/tree/main/examples/corda_finetuning}{examples/corda\_finetuning}.
\end{abstract}


\section{Introduction}
\label{introduction}

\IEEEPARstart{L}{arge} language models (LLMs) have demonstrated exceptional performance across a variety of challenging tasks, such as question answering \cite{devlin2018bert,radford2018improving}, machine translation \cite{brown2020language}, common-sense reasoning \cite{bosselut2019comet}, and instruction following \cite{ziegler2019fine}. 
Despite their impressive capabilities, adapting pre-trained LLMs to downstream tasks often entails substantial computational and memory demands due to their large size, posing significant challenges in real-world applications. 
To facilitate more resource-efficient adaptation, parameter-efficient fine-tuning (PEFT) techniques have been developed.
In contrast to fulll fine-tuning, 
these methods significantly reduce the number of trainable parameters during fine-tuning by 
optimizing only the introduced lightweight adapters~\cite{houlsby2019parameter,hu2022lora,he2022towards} or tokens~\cite{lester2021power,li2021prefix,razdaibiedina2023residual}, while freezing the pre-trained weights.

Particularly, low-rank adaptation (LoRA) \cite{hu2022lora} suggests that weight updates during LLM fine-tuning exhibit a low-rank structure, and accordingly employs two low-rank matrices with low intermediate dimensions as learnable adapters for each linear layer. 
LoRA has gained increasing popularity as its low-rank matrices can be incorporated into the original linear weights after fine-tuning, ensuring that the model architecture remains unchanged and incurs no additional inference overhead. 
Subsequent research has enhanced LoRA by using adaptive intermediate low ranks across layers \cite{zhang2023adaptive,valipour2022dylora,zhang2023increlora}, 
developing new learning schemes \cite{liu2024dora,lora+},
reducing the number of trainable parameters further \cite{kopiczko2023vera,renduchintala2023tied}, and combining it with quantization or pruning \cite{dettmers2024qlora,xu2024qalora,li2024loftq,zhang2023pruning}. 
Despite these efforts, the performance of LoRA typically falls short of full fine-tuning. Moreover, most existing studies are only concerned with the adaptation performance on downstream tasks, but ignore the problem of forgetting induced by fine-tuning. Even though LoRA only has a small number of learnable parameters, the resulting weight updates may still lead to catastrophic forgetting of the inherent world knowledge, which is acquired from massive pre-training data and can be difficult to recover once lost.

\begin{figure}[t]
	\centering
	\includegraphics[width=1.\linewidth]{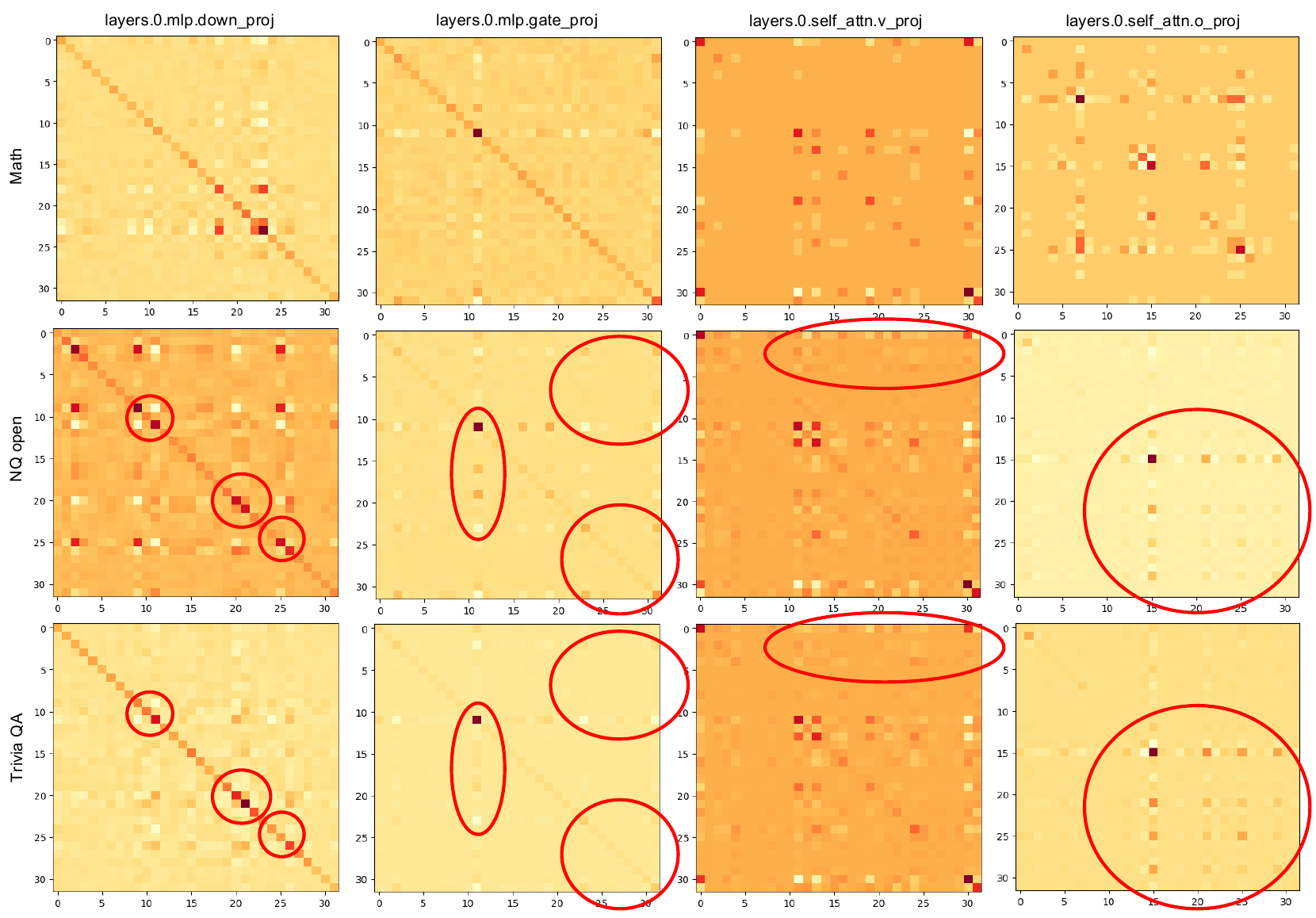} 
	\caption{Covariance matrix visualization for 4 different input activations in the 0-th block. We down-sample the heatmaps into $32\times 32$. Similar patterns are marked in red circles. More results are provided in the Appendix.}
	\label{cov_vis}
\end{figure}

We posit that these issues derive from the disregard for task awareness when initializing the LoRA adapters. The two low-rank matrices of LoRA adapters are respectively initialized by Gaussian initialization and zero, no matter what downstream task is to learn or what pre-trained knowledge needs to preserve. As a result, the initialized adapters are task-agnostic and may not be the optimal choice for the downstream task or preserving pre-trained knowledge. Some studies have proposed parameterization and initialization variants for LoRA adapters \cite{koohpayegani2024nola,li2024vblora,hayou2024impact,balazy2024lora,meng2024pissa}. However, none of these methods involve data context or task awareness in their process of building adapters. As shown in Fig.~\ref{cov_vis}, we observe that the covariance matrices of the input of linear layers share similar outlier distribution patterns for related tasks, \emph{i.e.,} NQ open \cite{lee-etal-2019-latent} and Trivia QA \cite{joshi-etal-2017-triviaqa}, while being discrepant from a different task \emph{i.e.,} Math \cite{yu2024metamath}. 
This indicates that the most responsive channels of weights vary with the kind of task triggered. 
Therefore, if it is possible to extract the weight components that are most correlated with the ability of a task, we can use them to initialize task-aware adapters for better adapting to this task, or freeze them to preserve their associated knowledge.

Inspired by this insight,
we present a context-oriented decomposition low-rank adaptation method (\textbf{CorDA}). 
Concretely, we sample a few data from the task of concern, such as questions from a QA dataset whose knowledge needs to be preserved, or instructions and responses of a fine-tuning dataset used to learn a new ability. We feed these samples into the pre-trained LLM and calculate the covariance matrices 
of each linear layer weight,  \emph{i.e.,} $C=XX^T\in\mathbb{R}^{d_{in}\times d_{in}}$, where $X$ is the input activation of this layer. Next, we apply singular value decomposition (SVD) to the product of each weight matrix $W\in\mathbb{R}^{d_{out}\times d_{in}}$ and its covariance matrix, expressed as $\verb|SVD|(WC)=U\Sigma V^T$, where the columns of $U$ and $V$ are singular vectors, and $\Sigma$ is a diagonal matrix containing singular values arranged in descending order. 
By doing so, the representative context captured by the covariance matrices can 
direct the orientation 
of decomposition, such that the largest several singular values and their singular vectors most correspond to the ability of this task. 
Thanks to the task awareness of our decomposition, we can choose to either freeze these task-specific principal components (knowledge-preserved adaptation mode, \emph{KPM}) or adapt them (instruction-previewed adaptation mode, \emph{IPM}). 
For KPM, we sample data from a QA dataset whose knowledge requires maintenance, and split the smallest $r$ singular values and vectors as learnable LoRA adapters, while freezing the remaining components to retain their associated knowledge. 
For IPM, we sample data from a fine-tuning dataset and use the top $r$ components as learnable adapters to better accommodate the new ability. 
After selecting weight components to initialize LoRA adapters, the inverse of $C$ is right-multiplied to ensure that the initial inference results remain unchanged. 
After fine-tuning, the learned adapter weights can also be merged into the frozen part, achieving the same efficiency as LoRA \cite{hu2022lora}.

Beyond CorDA \cite{yang2024corda}, which is published in NeurIPS 2024, we further propose \textbf{CorDA++} in this paper, offering substantial improvements and extended applicability. We notice that CorDA relies on a random sampling process to collect covariance matrices. While a small number of samples, usually $256$, suffices to perform our context-oriented decomposition, the resulting adapter initialization involves variability as 
the data for sampling may not be equally representative \cite{yang2022towards}. 
Consequently, adaptation results may become less effective when less-representative samples lacking critical task characteristics are selected. 
Additionally, the intermediate rank of adapters remains constant across layers in CorDA. 
However, certain layers may exhibit higher sensitivity to specific tasks, 
demanding more ranks to effectively encapsulate task-related abilities into principal components. 
Fine-tuning with adaptive ranks across layers helps to better allocate parameter budget and improve adaptation performance \cite{zhang2023adaptive}.

To this end, we first propose a metric, which is dependent on the singular value distributions of both $C$ and $WC$, to evaluate the compactness of task-specific principal components after our context-oriented decomposition. Minimizing this metric reduces the loss of task capabilities when truncating the decomposed components. 
Based on this metric, we then integrate two adaptive strategies into CorDA++, namely \emph{dynamic covariance selection} and \emph{dynamic rank allocation}. 
In dynamic covariance selection, we collect a pool of covariance matrix candidates for each linear layer through multiple rounds of data sampling. 
For each layer, we select the candidate with the lowest score to guide our decomposition. 
This method provides each layer with the most representative covariance matrix with relatively stronger contextual relevance from the candidates,
largely mitigating the variability caused by a single random sampling for all layers.

In dynamic rank allocation, we develop a progressive filtering process, where the decomposed components across all layers are iteratively filtered based on their compactness scores. 
In KPM, the filtered weight components are used as initialized adapters, while in IPM, the remaining weight components serve as adapters. 
This process continues until the total adapter parameter amount
aligns with the target parameter budget of low-rank adaptation. 
In this way, the learnable adapters across layers have dynamic ranks according to each layer's sensitivity to the specific task. 
As shown in Fig.~\ref{dynconv_dynr}, compared to the original decomposition method in CorDA, the two strategies in CorDA++ more effectively compact task-specific capabilities into the principal components, 
which can further improve the preservation of pre-trained knowledge in KPM and enhance fine-tuning performance in IPM.
In experiments, CorDA++ surpasses its original version in both modes, providing users with flexibility in choosing between stronger fine-tuning performance or better pre-trained knowledge preservation according to practical demands. 
For KPM, our method not only enjoys better downstream performances than LoRA on Math \cite{cobbe2021training,yu2024metamath}, Code \cite{chen2021evaluating,austin2021program}, and Instruction Following \cite{zheng2024judging}, but also largely alleviates 
the degradation of world knowledge, as evaluated on TriviaQA \cite{joshi-etal-2017-triviaqa}, NQ open \cite{lee-etal-2019-latent}, and WebQS \cite{berant2013semantic} benchmarks. 
For IPM, our method further enhances fine-tuning performance on downstream tasks, achieving faster convergence and surpassing strong baselines such as LoRA \cite{hu2022lora}, AdaLoRA \cite{zhang2023adaptive}, DoRA \cite{liu2024dora}, and PiSSA \cite{meng2024pissa}.

Compared to the conference version of NeurIPS 2024, we also extend the applicability of our method. We combine our CorDA++ with quantization by using full precision for the learnable adapters initialized by the principal components, while quantizing the frozen weight components with small singular values into Normal Float 4-bit \cite{dettmers2022bit}. 
Due to the heavy-head and long-tail singular value distribution of our context-oriented decomposition as illustrated in Fig.~\ref{eigen_dist}, the quantization of CorDA++ leads to 
less information loss and
outperforms existing quantization methods including QLoRA \cite{dettmers2024qlora}, LoftQ \cite{li2024loftq}, and QPiSSA \cite{meng2024pissa}. 
Moreover, we apply CorDA++ to vision language model fine-tuning and show that our KPM with visual tokens can also alleviate the loss of zero-shot visual question answering ability 
when fine-tuning LLaVA-1.5 \cite{liu2023visual,liu2024improved} on a new dataset.

\section{Related Work}
\label{related_work}

\textbf{Parameter-Efficient Fine-Tuning.}
As large language models (LLMs) scale up to tens or even hundreds of billions of parameters \cite{achiam2023gpt,bie2023renaissance}, full-parameter fine-tuning becomes prohibitively expensive in terms of both computation and memory overhead \cite{zhao2024galore,yangtowards}.
To alleviate resource constraints, parameter-efficient fine-tuning (PEFT) has emerged as a solution that updates only a small subset of parameters during adaptation \cite{ding2023parameter,xu2023parameter}.
One prominent category of PEFT techniques involves adapter-based approaches, where additional modules are inserted into the LLM and trained while keeping the original model frozen \cite{houlsby2019parameter,he2022towards,lei2023conditional,mahabadi2021parameter,pfeiffer2021adapterfusion,karimi2021compacter}.
Another stream of work employs soft prompts, which are either prepended to the input or injected into hidden layers as trainable vectors \cite{lester2021power,li2021prefix,razdaibiedina2023residual,zhu2023spt}.
Despite their efficiency, most of these techniques alter the model architecture or introduce extra inference overhead.
LoRA \cite{hu2022lora}, motivated by the low-rank nature of fine-tuned weight updates \cite{li2018measuring,aghajanyan2020intrinsic}, introduces two low-rank matrices into each linear layer, enabling adaptation without architectural changes or extra inference cost.
Various extensions of LoRA have been developed including adaptive ranks across layers \cite{zhang2023adaptive,valipour2022dylora,zhang2023increlora}, improving adapter design \cite{liu2024dora,chavan2023one,qiu2023controlling,zhao2024galore}, combining LoRA with model 
compression techniques \cite{zhang2023pruning,dettmers2024qlora,xu2024qalora,li2024loftq,guo2024lqlora}
and mixture-of-expert mechanisms \cite{liu2023moelora,dou2023loramoe}, and introducing alternative adapter initialization and parameterization schemes \cite{koohpayegani2024nola,li2024vblora,hayou2024impact,balazy2024lora,meng2024pissa,wang2024milora}. 
However, common limitations among existing LoRA-based approaches include the lack of data context when constructing adapters and the ignorance of pre-trained knowledge preservation.
Meanwhile, recent studies have demonstrated the value of data context in facilitating model quantization and compression \cite{lin2023awq,yuan2023asvd,lee2024owq}.
To incorporate data context into the initialization of LoRA adapters, we develop context-oriented decomposition in this paper, which injects task awareness into the principal components and thus supports both knowledge-preserved and instruction-previewed adaptation.

\textbf{Knowledge Forgetting.}
When deep learning models are adapted to new tasks, they often suffer from catastrophic forgetting, where previously learned knowledge is rapidly lost \cite{goodfellow2013empirical,rebuffi2017icarl,kirkpatrick2017overcoming,rao2019continual,lopez2017gradient,yang2023neural-iclr,yang2023neural,liu2024enhancing}.
To address this issue, various strategies have been explored based on knowledge distillation \cite{li2017learning,hou2019learning}, rehearsal \cite{riemer2018learning,yang2023neural-iclr}, and dynamic architecture \cite{yan2021dynamically}.
With the rise of large-scale models \cite{wu2024continual}, preserving pre-trained world knowledge, which is acquired through extensive pre-training on massive corpora, has become increasingly important and challenging~\cite{he2023continual,zhai2023investigating,scialom-etal-2022-fine,gupta2023continual,ibrahim2024simple}.
Some studies have proposed freezing pre-trained layers while introducing auxiliary structures such as additional LLaMA blocks \cite{wu2024llama} or mixture-of-expert modules \cite{dou2023loramoe}, to balance knowledge retention and task adaptation.
Another line of work focuses on merging techniques to integrate capabilities from multiple models  \cite{zhang2023composing,yu2023language,zhu2024model}.
In contrast to these approaches, our method preserves world knowledge within the framework of low-rank adaptation, without altering the model architecture or relying on any post-processing.

\textbf{Low-rank Decomposition for LLMs.}
Due to the low-rank structure inherent in LLM weight matrices, singular value decomposition (SVD) as a fundamental tool for identifying the most informative directions in the parameter space, has been widely leveraged in the areas of low-rank adaptation \cite{zhang2023adaptive,meng2024pissa,zhao2024galore}, model compression \cite{hsu2022language,yuan2023asvd,wang2025svdllm,li2025optimizing}, post-training quantization \cite{yao2024exploring,li2025svdquant}, and fine-tuning with quantization \cite{guo2024lqlora,li2024loftq,meng2024pissa}.
Beyond the reconstruction loss of SVD, the outlier distribution of activations is another important factor that influences how much the decomposed components contribute to a task-specific performance. 
To this end, some studies combine extra information when performing SVD, 
such as Fisher information in FWSVD \cite{hsu2022language}, activation magnitude in ASVD \cite{yuan2023asvd}, and data whitening in SVD-LLM \cite{wang2025svdllm}.
However, for low-rank adaptation, data context has rarely been considered.
PiSSA \cite{meng2024pissa} uses the plain SVD for linear weights and extracts the principal components to initialize LoRA adapters. AdaLoRA \cite{zhang2023adaptive} adaptively allocates parameter budget by formulating the incremental matrices in the form of SVD. 
Our work differs from these methods in that we utilize covariance matrices that better capture outlier distribution to orientate the decomposition, such that the resulting principal components can be associated with the task of concern.

\section{Low-rank Adaptation and Context-oriented Singular Value Decomposition}
\label{methods}

In this section, we first review low-rank adaptation (LoRA) in Sec. \ref{lora}, and then present our context-oriented decomposition (CO-SVD) in Sec. \ref{decomposition}. Finally, in Sec. \ref{dist_analysis}, we analyze the singular value distribution of CO-SVD to demonstrate its benefits and potential for enhancing LoRA.

\begin{figure*}[t]
	\centering
	\includegraphics[width=1.\linewidth]{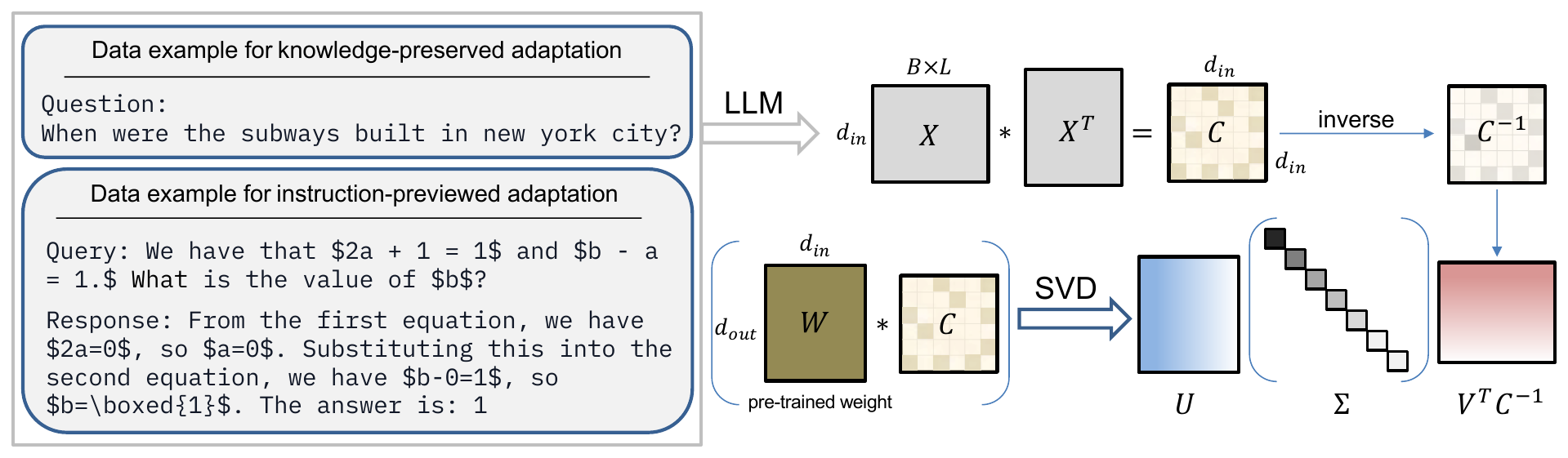}
	\caption{An illustration of our proposed 
		context-oriented SVD, which leverages data context from the target task and performs SVD on the product of weight matrix and its corresponding covariance matrix.
		$C^{-1}$ is right-multiplied to reconstruct $W$. 
	}
	\label{fig1}
\end{figure*}

\begin{figure*}[t]
	\centering
	\begin{subfigure}{0.2455\textwidth}
		\centering
		\includegraphics[width=1.\linewidth]{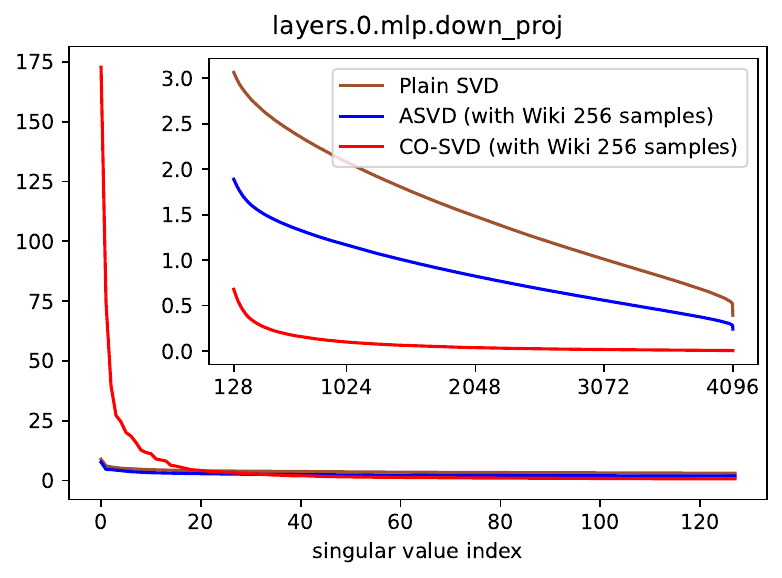}
		\label{eigen_1}
	\end{subfigure}
	\begin{subfigure}{0.2455\textwidth}
		\centering
		\includegraphics[width=1.\linewidth]{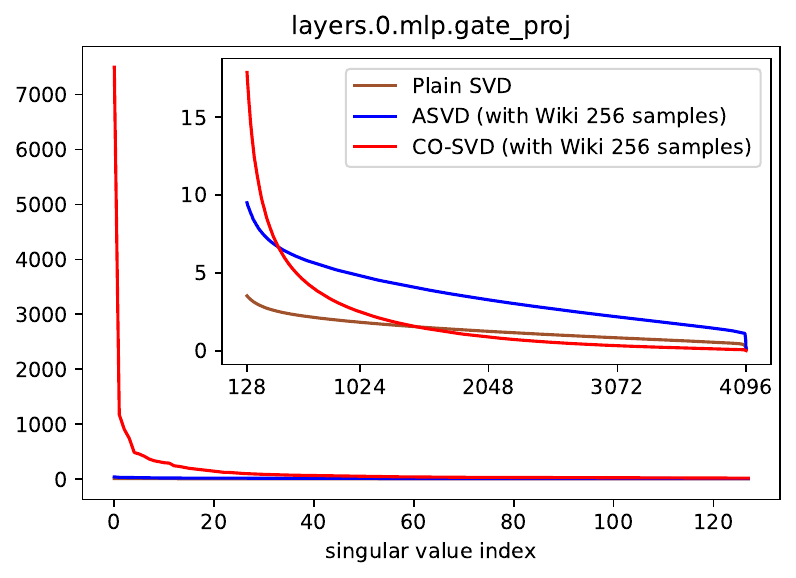}
		\label{eigen_2}
	\end{subfigure}
	\begin{subfigure}{0.2455\textwidth}
		\centering
		\includegraphics[width=1.\linewidth]{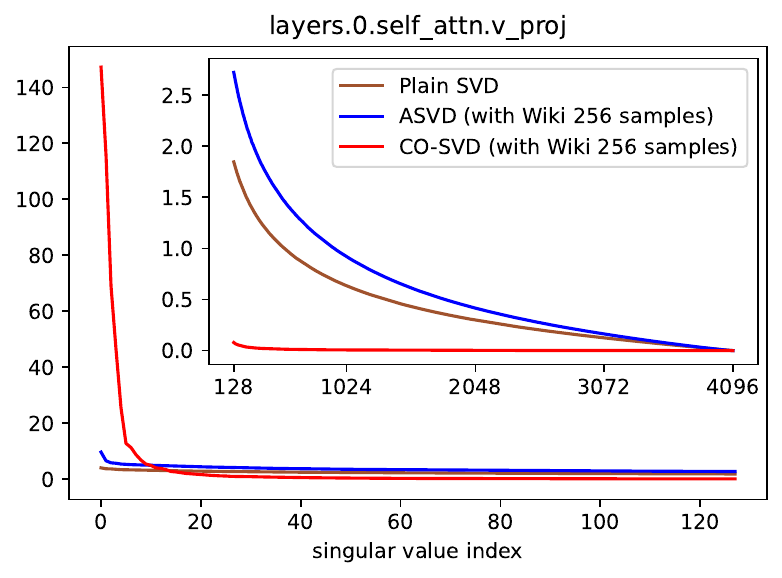}
		\label{eigen_3}
	\end{subfigure}	
	\begin{subfigure}{0.2455\textwidth}
		\centering
		\includegraphics[width=1.\linewidth]{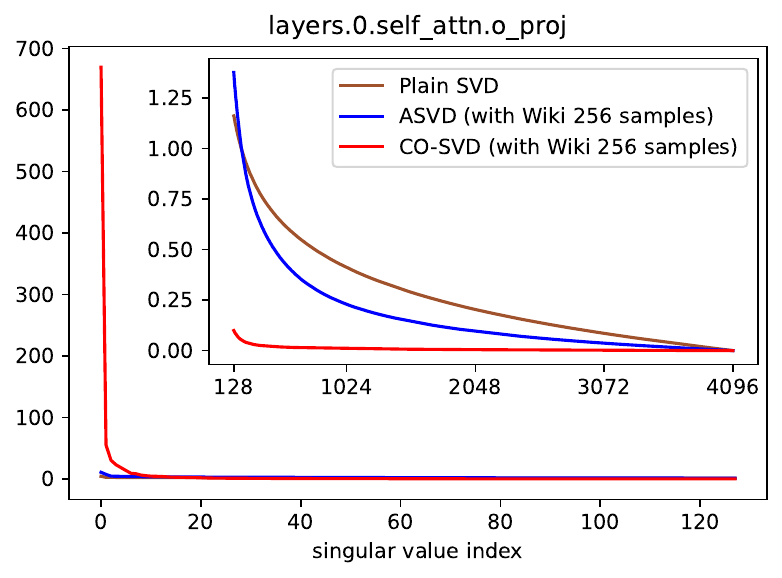}
		\label{eigen_4}
	\end{subfigure}
	\caption{Singular value distribution of plain SVD, ASVD, and our proposed context-oriented SVD (CO-SVD). The embedded figures show singular values with indices from 128 to 4096, and main figures illustrate the top 128 singular values. 
		}
	\label{eigen_dist}
\end{figure*}

\subsection{Low-Rank Adaptation}
\label{lora}

LoRA observes that the weight updates after LLM fine-tuning exhibit an inherently low-rank structure \cite{hu2022lora}.
Based on this insight, it approximates the weight change using two learnable low-rank matrices while keeping the pre-trained weights frozen throughout the fine-tuning process. 
Denote a pre-trained weight as $W\in\mathbb{R}^{d_{out}\times d_{in}}$. The updated weight after fine-tuning can be expressed as:
\begin{equation}
	W^*=W+\Delta W = W + {B} {A},
\end{equation}
where  $W^*$ is the fine-tuned weight, $\Delta W$ refers to the weight change, and ${B}\in\mathbb{R}^{d_{out}\times r}$, ${A}\in\mathbb{R}^{r\times d_{in}}$ are the learnable low-rank matrices with a small intrinsic rank of $r\ll \min{(d_{out}, d_{in})}$. 
To avoid deviating the inference result from the pre-trained model, LoRA uses the Kaiming initialization \cite{he2015delving} for $A$, and initializes $B$ by an all-0 matrix such that $\Delta W=0$ at the start of fine-tuning. 
As only the low-rank matrices $A$ and $B$ are learnable in fine-tuning, the number of trainable parameters is significantly reduced. 
After fine-tuning, the product $BA$ can be incorporated into the pre-trained weight $W$ without altering model architecture and incurring any additional inference cost. 

While LoRA improves training efficiency, its adapter initialization is ignorant of any data context related to specific tasks. As a result, the adaptation performance lags behind full fine-tuning, and often leads to pre-trained knowledge forgetting.

\begin{figure*}[t]
	\centering
	\begin{subfigure}{0.495\textwidth}
		\centering
		\includegraphics[width=0.8\linewidth]{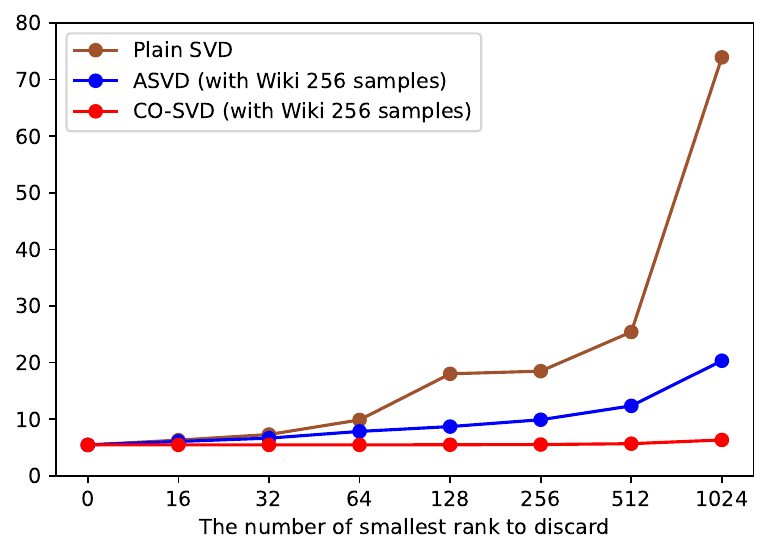}
		\vspace{-2mm}  
		\caption{Perplexity on Wikitext-2}
		\label{fig:wiki}
	\end{subfigure}
	\begin{subfigure}{0.495\textwidth}
		\centering
		\includegraphics[width=0.8\linewidth]{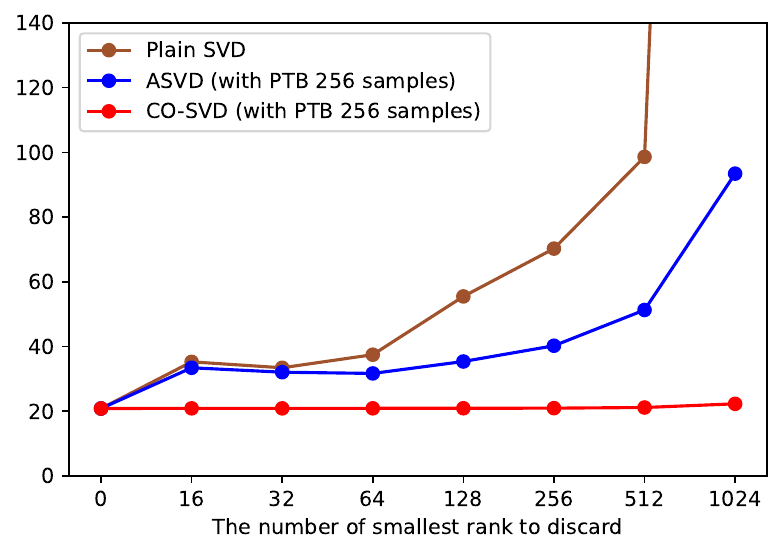}  
		\vspace{-2mm}
		\caption{Perplexity on PTB}
		\label{fig:ptb}
	\end{subfigure}	
	\caption{Perplexity (lower is better) on (a) Wikitext-2 and (b) Penn TreeBank (PTB) after decomposing each weight matrix in LLaMA-2-7B and discarding the smallest $r$ singular values and their singular vectors. 
		Compared to plain SVD and ASVD, our context-oriented SVD (CO-SVD) incurs only a slight performance drop even when removing the bottom 1024 components. 
		}
	\label{fig2}
\end{figure*}

\begin{figure*}[t]
	\centering
	\begin{subfigure}{0.495\textwidth}
		\centering
		\includegraphics[width=0.8\linewidth]{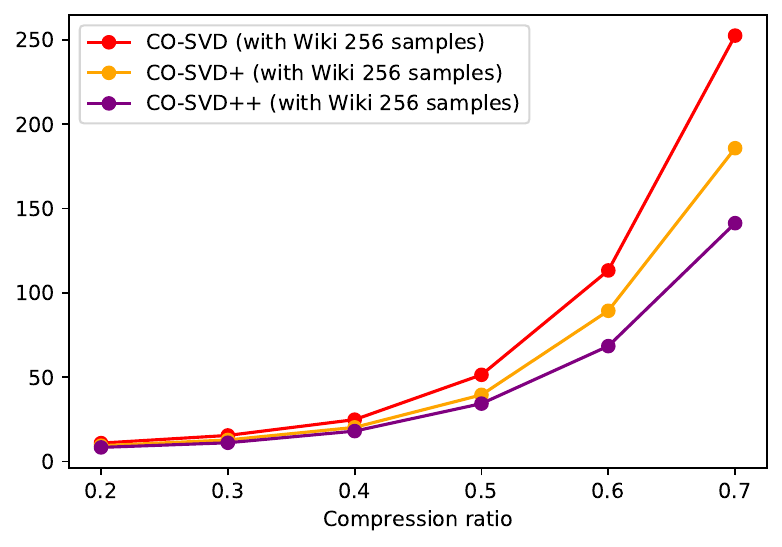}
		\vspace{-2mm}  
		\caption{Perplexity on Wikitext-2}
		\label{dynr_dynconv_wiki}
	\end{subfigure}
	\begin{subfigure}{0.495\textwidth}
		\centering
		\includegraphics[width=0.8\linewidth]{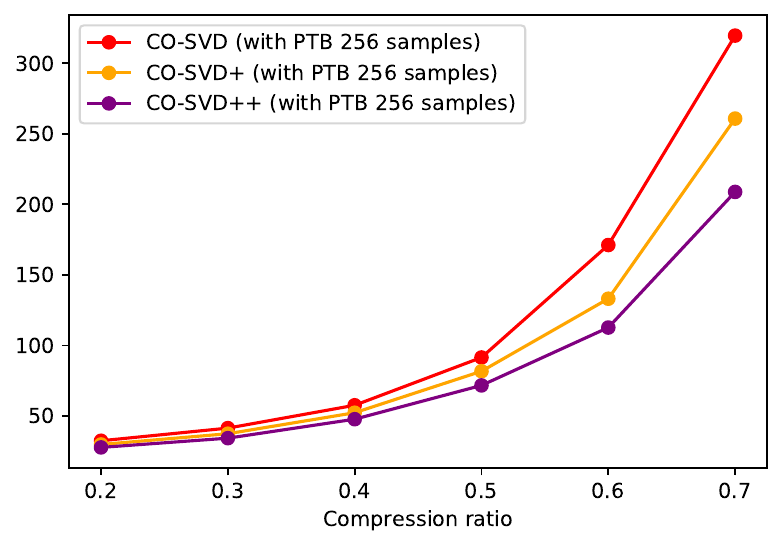}  
		\vspace{-2mm}
		\caption{Perplexity on PTB}
		\label{dynr_dynconv_ptb}
	\end{subfigure}	
	\caption{Perplexity at different compression ratios by progressively removing bottom components for the original CO-SVD, CO-SVD+ (with dynamic rank allocation), and CO-SVD++ (with both dynamic rank allocation and dynamic covariance selection).}
	\label{dynconv_dynr}
\end{figure*}

\subsection{Context-Oriented Singular Value Decomposition}
\label{decomposition}

Pre-trained large language models inherently possess multi-faced capabilities. Although the input prompts from different tasks are processed through the same pre-trained weights, they activate distinct functionalities within the model, which means that the most responsive weight components vary with the task triggered. 
These variations manifest in the activation patterns of each layer, whose covariance matrices can reveal task-specific outlier structures. As shown in Fig. \ref{cov_vis}, covariance matrices from related tasks share similar outlier distribution patterns. 
Therefore, the covariance matrix encodes rich contextual information about the task. Inspired by the insight, we propose to utilize covariance matrices to extract the weight components aligned with a specific task. 

The process of our context-oriented decomposition (CO-SVD) is shown in Fig. \ref{fig1}.
First, we randomly sample a small batch of training data from the target task, such as questions of a QA task or queries and responses of a Math dataset, and feed it into the pre-trained LLM. Let $X\in\mathbb{R}^{d_{in} \times B L}$ represent the input activation of a linear layer, where $d_{in}$ denotes the input dimension, $B$ is the number of samples, and $L$ is the sequence length. The corresponding covariance matrix is computed as  $C=XX^T\in\mathbb{R}^{d_{in}\times d_{in}}$. Next, we perform SVD on the product of the weight matrix and its covariance matrix as follows:
\begin{equation}
	\verb|SVD|(WC)= U\Sigma V^T=\sum_{i=1}^{R} \sigma_i \mathbf{u}_i \mathbf{v}_i^T,
\end{equation}
where $W\in\mathbb{R}^{d_{out} \times d_{in}}$ is the weight matrix. $U\in\mathbb{R}^{d_{out} \times d_{out}}$ and $V\in\mathbb{R}^{d_{in} \times d_{in}}$ are orthogonal and consist of the left and right singular vectors, denoted by $\mathbf{u}_i \in \mathbb{R}^{d_{out}}$ and $\mathbf{v}_i \in \mathbb{R}^{d_{in}}$, respectively. The diagonal matrix $\Sigma\in\mathbb{R}^{d_{out} \times d_{in}}$ contains the singular values $\sigma_i$ arranged in descending order. $R$ corresponds to the number of non-zero singular values, \emph{i.e.,} the rank of $WC$, satisfying $R\le \min\{d_{out}, d_{in}\}$.

As in LoRA, it is necessary to preserve the original inference behavior of the pre-trained model. Therefore, we reconstruct the weight matrix by: 
\begin{equation}
	\hat{W} = \verb|SVD|(WC)C^{-1}=U\Sigma (V^T C^{-1})=\sum_{i=1}^{R} \sigma_i \mathbf{u}_i \mathbf{\hat{v}}_i^T,
	\label{decomposed-reconstruct}
\end{equation}
where $C^{-1}$ represents the inverse of $C$, and $\hat{\mathbf{v}}_i^T$ denotes the $i$-th row vector of $V^T C^{-1}$. 
When $C$ is not invertible, we employ a regularization strategy that ensures invertibility by dynamically adjusting its diagonal. Specifically, we compute the average of the diagonal entries in $C$, multiply it by a positive coefficient, and add the result to the diagonal elements.
We then calculate the  $\ell_2$ distance between $CC^{-1}$ and an identity matrix. If it exceeds a pre-defined threshold, the coefficient is doubled and the process is repeated until the distance falls below the threshold.

\subsection{Singular Value Distribution Analysis}
\label{dist_analysis}

In this subsection, we analyze our proposed CO-SVD by examining its singular value distribution and its advantage in compacting task-specific abilities into the principal components.

As shown in Fig. \ref{eigen_dist}, we visualize the singular values $\sigma_i $ of the plain SVD (directly decomposing the weight matrix), ASVD \cite{yuan2023asvd}, and our CO-SVD for 4 different linear layers in the first block of LLaMA-2-7B. 
The results indicate that our method exhibits a heavy head, characterized by several extremely large singular values concentrated in the top components. This phenomenon may be attributed to the outliers from data activations whose covariance matrices are involved in our decomposition. 
Our method also displays a long-tail distribution. The singular values of CO-SVD since the 128-th one are significantly smaller and closer to $0$ than plain SVD and ASVD. 
This distribution behavior potentially brings the benefit of compacting the task-specific abilities into the leading principal components. More visualization results of other layers are provided in the Appendix. 

To further validate such benefit, we perform different decomposition methods, discard the smallest $r$ singular values and their singular vectors, and test the model performance using the remaining weight components. As shown in Fig. \ref{fig2}, when $r$ ranges from 0 to 1024, the performance of plain SVD is getting worse rapidly. ASVD considers data context by incorporating activation absolute means into decomposition, and helps to relieve the deterioration compared to plain SVD. But the perplexity of ASVD also diverges steeply when $r$ is higher than 128. 
In contrast, the model performance of CO-SVD only drops slightly on both Wikitext-2 \cite{merity2016pointer} and PTB \cite{marcus1993building}, even when $r$ reaches 1024. 
We also test the effect of sample number and dataset choice in the Appendix, where we show that CO-SVD with 32 samples, despite performing marginally worse than using 256 samples, still works well in maintaining a stable performance when truncating the decomposed components. Besides, we observe that collecting covariance matrices from the same dataset as the one for inference achieves better performance, which indicates that capturing the correct data context in CO-SVD helps to associate the principal components with the target task. 

These results imply that CO-SVD is capable of aggregating task-specific abilities effectively into its principal components by collecting data context from a few samples and guiding the decomposition with their covariance matrices. 
The bottom components correspond to the redundant part regarding the target task, and thus removing them causes minor performance drop. 
Accordingly, when applying CO-SVD for LoRA adapter initialization, we can decide to either freeze the principal components for preserving their associated task-specific ability, or adapt them for better accommodating a new task, which leads to our two implementation modes, knowledge-preserved adaptation and instruction-previewed adaptation, as we will detail in Sec. \ref{sec_modes}.

\section{Dynamic Context-oriented Decomposition}

Based on our proposed context-oriented decomposition, we further enhance our method by analyzing the compactness of the task-specific principal components in Sec. \ref{metric}, and proposing the dynamic covariance selection and dynamic rank allocation strategies in Sec. \ref{dyn_conv} and Sec. \ref{dyn_rank}, respectively.

\subsection{Compactness Metric}
\label{metric}

Given a linear layer with input activation $X\in\mathbb{R}^{d_{in} \times B L}$ and weight matrix $W\in\mathbb{R}^{d_{out}\times d_{in}}$, when truncating the weight components using our context-oriented decomposition, we denote the truncated weight as $W'$, and $\Delta W = W'-W$.
Then the output activation shifting caused by truncation can be expressed as $\Delta Y=\Delta W X$. We aim to minimize the maximal absolute shift across all tokens, then we have:
\begin{equation}\label{eq_deltay}
	\left\| \Delta Y \right\|_1 \le \sqrt{d_{out}} \left\| \Delta Y \right\|_2 \le \sqrt{d_{out}} \left\| \Delta W \right\|_2 \left\| X \right\|_2,
\end{equation}
where 
\begin{equation}
    \left\| \Delta Y \right\|_1=\max_{1\le j \le BL}\sum_{i=1}^{d_{out}}\left| \Delta Y_{ij} \right|, \notag
\end{equation}
and $\left\| A \right\|_2$ denotes the spectral norm of a matrix $A$, \emph{i.e.,} its maximal singular value. 
Considering $C=XX^T$ is a symmetric positive semi-definite matrix, we have $ \left\| X \right\|_2 =\sigma_{max}(X)=\sqrt{\sigma_{max}(C)}$, where $\sigma_{max}$ denotes the maximal singular value. 
In our context-oriented decomposition, we have
\begin{equation}
	\Delta W = U_{[:,-r:]}\Sigma_{[-r:]}(V^T)_{[-r:,:]}C^{-1}, \notag
\end{equation}
where $r$ is the truncated rank, \emph{i.e.,} discarding the bottom $r$ components. $U_{[:,-r:]}$ represents the last $r$ column vectors of $U$, $\Sigma_{[-r:]}$ refers to the last $r$ diagonal elements of $\Sigma$, and $(V^T)_{[-r:,:]}$ denotes the last $r$ row vectors of $V^T$. 
Then we have:
\begin{align}\label{eq_deltaw}
	 \left\| \Delta W \right\|_2 & =  \left\| U_{[:,-r:]}\Sigma_{[-r:]}(V^T)_{[-r:,:]}C^{-1} \right\|_2 \notag \\
	 &\le \left\| U_{[:,-r:]}\Sigma_{[-r:]}(V^T)_{[-r:,:]}\right\|_2  \left\|C^{-1} \right\|_2 \\
	 &=    \frac{\sigma_{-r}}{\sigma_{min}(C)}, \notag
\end{align}
where $\sigma_{-r}$ is the $r$-th singular value \emph{from the end} in our CO-SVD that decomposes $WC$, \emph{i.e.,} the largest singular value of $U_{[:,-r:]}\Sigma_{[-r:]}(V^T)_{[-r:,:]}$, and $\sigma_{min}(C)$ is the minimal singular value of $C$, satisfying $1/{\sigma_{min}(C)}=\left\|C^{-1} \right\|_2$.

Combining Eq. (\ref{eq_deltay}) and Eq. (\ref{eq_deltaw}), we have:
\begin{equation}\label{score}
	\left\| \Delta Y \right\|_1 \le \frac{\sqrt{d_{out}\sigma_{max}(C)}}{\sigma_{min}(C)} \cdot \sigma_{-r}=\pi(C) \cdot \sigma_{-r},
\end{equation}
where we define $\pi(C)\vcentcolon={\sqrt{d_{out}\sigma_{max}(C)}}/{\sigma_{min}(C)}$.
It indicates that the maximal absolute shift caused by truncating the smallest $r$ ranks is upper bounded by a metric that depends on $\pi(C)$ and the singular value distribution of $WC$. 
When the metric is large, the model is prone to lose much model performance when truncating the last $r$ components of CO-SVD. Conversely, a small value implies that the task-specific abilities are compact in the principal components. 

Built upon this metric, we develop two strategies in the following subsections. 

\subsection{Dynamic Covariance Selection}
\label{dyn_conv}
In CO-SVD, if we randomly sample data to collect covariance matrices, variability will be introduced as samples are not equally representative. As Eq. (\ref{score}) indicates, covariance matrix selection directly influences the compactness. When less-representative samples are selected, the resulting principal components may not fully capture the characteristics of the target task, and the adaptation results may be less effective. 

To this end, we propose dynamic covariance selection. We perform multiple rounds of data sampling to get $\mathcal{D}=\{\mathcal{I}_1, \mathcal{I}_2,...,\mathcal{I}_N\}$, where $\mathcal{I}_i$ is the $i$-th batch and $N$ denotes the number of sampling rounds. We feed each batch into the pre-trained LLM, and get its corresponding covariance matrix $C^{(l)}_i$ for each linear layer, where $l$ is the layer index. Next, we calculate the covariance selection score as follows:
\begin{equation}\label{cov_score}
	s(C_i^{(l)}) =\log \left( \pi(C_i^{(l)}) \right) \sum_{r=1}^{R} \frac{\sigma_{-r}}{\sigma_{max}},
\end{equation}
where we apply a logarithmic transformation to $\pi(C_i^{(l)})$ to compress its scale and facilitate ranking. We normalize $\sigma_{-r}$ by dividing them by their maximal value $\sigma_{max}$ and compute the sum to evaluate the overall spectral concentration. A smaller value indicates that the spectral energy is concentrated in the top components, suggesting stronger compactness.

For each layer, we select the covariance matrix with the lowest covariance selection score from the candidates, \emph{i.e.,}
\begin{equation}\label{select_C}
	C^{(l)} = \argmin s(C_i^{(l)}),\quad i=1,2...,N.
\end{equation}
In this way, each layer is provided with a relatively more representative covariance matrix for CO-SVD, mitigating the variability from a single random sampling shared by all layers. 
The process of dynamic covariance selection is summarized in Algorithm \ref{alg1},

\begin{algorithm}[t]
	\small
	\caption{
		Dynamic covariance selection
	}
	\label{alg1}
	\begin{algorithmic}[1]
		\REQUIRE Pre-trained weight matrices for all layers $\{W^{(l)}\}_{l=1}^L$, multiple rounds of data sampling $\mathcal{D}=\{\mathcal{I}_i\}_{i=1}^N$.    
		\STATE Calculate covariance matrices for all layers from each sampling, $C_i^{(l)}$, $1\le i \le N$, $1\le l \le L$, by feeding each $\mathcal{I}_i$ into the LLM
		\FOR{$l=1$ to $L$}
		\FOR{$i=1$ to $N$}
		\STATE Apply SVD to $C_i^{(l)}$ to calculate $\pi(C_i^{(l)})$
		\STATE Apply CO-SVD with $W^{(l)}$ and $C_i^{(l)}$ to get the distribution of singular values $\sigma$
		\STATE Calculate $s(C_i^{(l)})$ by Eq. (\ref{cov_score}) 
		\ENDFOR
		\STATE Get $C^{(l)}$ by Eq. (\ref{select_C}) 
		\ENDFOR
		\STATE Output  $\{C^{(l)}\}_{l=1}^L$ 
	\end{algorithmic}
\end{algorithm}

\subsection{Dynamic Rank Allocation}
\label{dyn_rank}

For a specific task, some layers in a pre-trained LLM may be more responsive, requiring more ranks to fully encapsulate its ability into the principal components. 
Dynamically allocating ranks across layers based on their spectral compactness allows the model to assign fewer learnable parameters to the adapters of layers that are less task-sensitive, utilizing the parameter budget more efficiently. 

Therefore, we propose dynamic rank allocation based on the same metric in Eq. (\ref{score}). 
After acquiring the covariance matrix of each linear layer, we perform our CO-SVD for all these layers, and introduce a progressive filtering process. Concretely, we maintain a truncating position list, $\{r^{(l)}\}_{l=1}^L$, $1\le r^{(l)}\le R^{(l)}$, where $L$ is the number of linear layers, $R^{(l)}$ denotes the full rank of the $l$-th layer's decomposition, and $r^{(l)}$ represents the number of ranks to filter for layer $l$. 
At the beginning of the process, we have $r^{(l)}=1, \forall 1\le l\le L$. Next, for each layer, we calculate the rank allocation score as follows:
\begin{equation}\label{rank_score}
	s^{(l)} = \log \left( \pi(C^{(l)}) \right)\frac{\sigma_{-r^{(l)}}}{ \sum_{k=1}^{R^{(l)}-r^{(l)}}  \sigma_{k}},
\end{equation}
where we keep the logarithmic term in Eq. (\ref{cov_score}), and divide $\sigma_{-r^{(l)}}$ by the cumulative sum of its preceding singular values to reflect its incremental contribution relative to prior components. 
A smaller ratio suggests that $\sigma_{-r^{(l)}}$ lies in the spectral tail with diminished importance to the target ability. 
We then filter one more rank for the layer that has the lowest $s^{(l)}$, \emph{i.e.,}
\begin{equation}\label{rank_allocate}
	r^{(l_0)} \leftarrow r^{(l_0)} + 1,\quad l_0=\argmin_l s^{(l)},\quad 1\le l \le L.
\end{equation} 
For knowledge-preserved adaptation, we use the filtered components to initialize LoRA adapters, while for instruction-previewed adaptation, we use the remaining components as the initialized adapters. The filtering process iteratively continues until the number of adapter parameters reaches the budget. 
The process of dynamic rank allocation is summarized in Algorithm \ref{alg2}.

As shown in Fig. \ref{dynconv_dynr}, we arm our CO-SVD with the two proposed strategies. 
CO-SVD with dynamic rank allocation (CO-SVD+) discards the truncated components by running Eq. (\ref{rank_allocate}) progressively until the remaining parameter number satisfies the compression ratio. It leads to a lower performance drop compared to the original CO-SVD that filters a fixed number of ranks across layers. 
Moreover, CO-SVD with both strategies (CO-SVD++) mitigates the ability loss of the target task to a larger degree.  
These results indicate that the two strategies can further enhance the compactness of task-specific ability within the principal components.

\begin{algorithm}[t]
	\small
	\caption{
		Dynamic rank allocation
	}
	\label{alg2}
	\begin{algorithmic}[1]
		\REQUIRE Pre-trained weight matrices for all layers $\{W^{(l)}\}_{l=1}^L$, the covariance matrices $\{C^{(l)}\}_{l=1}^L$, the initial truncating position list $\{r^{(l)}\}_{l=1}^L$, $r^{(l)}=1, \forall 1\le l\le L$, the budget of adapter parameters $\tau$, the adaptation mode: knowledge-preserved mode (KPM) or instruction-previewed mode (IPM), the current number of adapter parameters $\tau'=0$.
		\FOR{$l=1$ to $L$}
		\STATE Apply SVD to $C^{(l)}$ to calculate $\pi(C^{(l)})$
		\STATE Apply CO-SVD with $W^{(l)}$ and $C^{(l)}$ to get the distribution of singular values $\sigma$
		\ENDFOR
		\WHILE{{True}}
		\STATE Calculate $s^{(l)}, \forall 1\le l \le L$ by Eq. (\ref{rank_score})
		\STATE Get the layer index  $l_0$, such that $l_0=\argmin_l s^{(l)}$    
		\STATE $r^{(l_0)} \leftarrow r^{(l_0)} + 1$
		\IF{KPM}
		\STATE {\color{gray} \# filtered components are initialized adapters in KPM}
		\STATE $\tau'=\sum_{l=1}^{L} (d_{in}^{(l)} + d_{out}^{(l)}) * r^{(l)}$
		\IF{$\tau'>\tau$}
		\STATE Break
		\ENDIF
		\ELSIF{IPM}
	    \STATE {\color{gray} \# remaining components are initialized adapters in IPM}
		\STATE $\tau'=\sum_{l=1}^{L} (d_{in}^{(l)} + d_{out}^{(l)}) * (R^{(l)} - r^{(l)})$
		\IF{$\tau'<\tau$}
		\STATE Break
		\ENDIF
		\ENDIF
		\ENDWHILE
		\STATE Output  $\{r^{(l)}\}_{l=1}^L$, initialized adapters, and the residual weights. 
	\end{algorithmic}
\end{algorithm}

\section{Adaptation Modes and Applications}
\label{sec_modes}

In this section, we introduce the two adaptation modes, knowledge-preserved adaptation and instruction-previewed adaptation, in Sec. \ref{KPA} and Sec. \ref{IPA}, respectively. We also apply our method to adaptation with quantization in Sec. \ref{QA}, and vision language models in Sec. \ref{VLMA}.

\begin{figure*}[t]
	\centering
	\includegraphics[width=1.\linewidth]{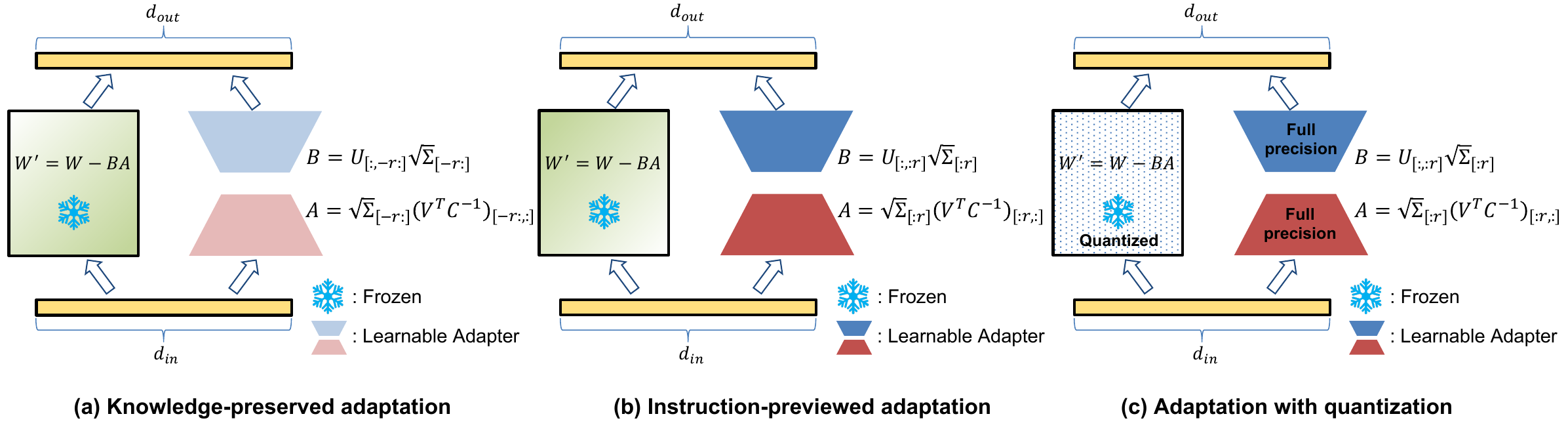}
	\caption{In (a) knowledge-preserved adaptation mode, we use the bottom $r$ components to initialize adapters with leading components frozen to preserve their associated pre-trained knowledge. In (b) instruction-previewed adaptation mode, we adapt the principal components to accelerate learning as they align with the most responsive directions to the new task in the decomposed subspace. In (c) adaptation with quantization, we quantize the residual weight and keep adapters in full precision. 
		}
	\label{adaptation_modes}
\end{figure*}

\subsection{Less Forgetting: Knowledge-Preserved Adaptation}
\label{KPA}

The knowledge-preserved adaptation mode (KPM) is developed for maintaining certain pre-trained world knowledge when fine-tuning on a new task. 
In this mode, we collect covariance matrices by sampling questions from question-answering (QA) datasets that cover the knowledge we aim to preserve, such as TriviaQA \cite{joshi-etal-2017-triviaqa} and Natural Questions \cite{kwiatkowski2019natural,lee-etal-2019-latent}. 
The principal components of our CO-SVD will capture the model's ability to retrieve the relevant knowledge.
As shown in Fig. \ref{adaptation_modes}-(a), we use the bottom $r$ components, \emph{i.e.,} the smallest $r$ singular values and their singular vectors in Eq. (\ref{decomposed-reconstruct}), to initialize learnable adapters as follows:
\begin{equation}
	B=U_{[:,-r:]} \sqrt{\Sigma}_{[-r:]}, \quad A = \sqrt{\Sigma}_{[-r:]} (V^T C^{-1})_{[-r:,:]},\label{KPA_eq}
\end{equation}
where $B\in\mathbb{R}^{d_{out}\times r}$, $A\in\mathbb{R}^{r \times d_{in}}$ represent the initialized low-rank adapters, and $\sqrt{\Sigma}_{[-r:]}$ is the square root of the smallest $r$ singular values in a diagonal matrix. 
The remaining components,
\begin{equation}\label{residual}
	W'=W-BA,
\end{equation}
are frozen during fine-tuning as a residual weight because they mainly correspond to the target QA ability to preserve. We calculate $W'$ as the difference between $W$ and $BA$ instead of summing the largest $R-r$ components to avoid numerical error from the decomposition and inversion operations.
When fine-tuning is completed, the learned adapters can be integrated into the residual weight as $W^* = W' + B^*A^*$, in the same way as LoRA. 
Our method in this mode can also achieve better fine-tuning performance on downstream tasks than LoRA, as verified by our experimental results.

\subsection{Faster Convergence: Instruction-Previewed Adaptation}
\label{IPA}

The instruction-previewed adaptation mode (IPM) will be preferable when the aim of adaptation is only to maximize performance on the downstream task, with no concern for knowledge retention. 
In this mode, we sample instructions and responses from the training dataset used for fine-tuning, such as the query to solve a math problem and its answer as shown in Fig. \ref{fig1}, to collect covariance matrices. 
In this way, the principal components of CO-SVD correspond to the most task-responsive part of the pre-trained weights. 
As shown in Fig. \ref{adaptation_modes}-(b), we use the leading $r$ components, \emph{i.e.,} the largest $r$ singular values and their singular vectors in Eq. (\ref{decomposed-reconstruct}), to initialize learnable adapters as follows:
\begin{equation}
	B=U_{[:,:r]} \sqrt{\Sigma}_{[:r]}, \quad A = \sqrt{\Sigma}_{[:r]} (V^T C^{-1})_{[:r,:]},\label{IPA_eq}
\end{equation}
where the subscripts $[:,:r]$ and $[:r,:]$ denote the first $r$ column vectors and the first $r$ row vectors of a matrix, respectively. 
$\sqrt{\Sigma}_{[:r]}$ is the square root of the largest $r$ singular values in a diagonal matrix.
The residual weight is still calculated by Eq. (\ref{residual}) and keeps frozen during fine-tuning. 
This mode allows the initialized adapters to capture the core characteristics of the target task in advance, facilitating faster convergence in fine-tuning. 
While a recent work \cite{meng2024pissa} also applies SVD to weights and selects the top $r$ components to initialize adapters, the plain SVD used in their work cannot leverage data context, and its compactness of task-specific ability is significantly inferior to CO-SVD, as demonstrated in Fig. \ref{eigen_dist} and Fig. \ref{fig2}. Accordingly, our method in IPM leads to better fine-tuning performance than multiple strong baselines in the experiments.

\subsection{Adaptation with Quantization}
\label{QA}

Quantization is an effective technique to reduce memory and computational consumption for LLM inference \cite{dettmers2022gpt3,xiao2023smoothquant}. It has also been applied to low-rank adaptation, to further improve training efficiency. For example, QLoRA \cite{dettmers2024qlora} quantizes the pre-trained weights to Normal Float 4-bit (NF4), while keeping the learnable adapters $A$ and $B$ in full precision. The following studies \cite{li2024loftq,meng2024pissa} adopt a similar strategy, applying quantization only to the frozen residual weights without changing adapters, but they differ in the initialization of adapters.

We also combine quantization with our adaptation method. As shown in Fig. \ref{adaptation_modes}-(c), the adapters initialized by the principal components of our CO-SVD are in full precision, while the residual weight is quantized. 
Thanks to the heavy-head and long-tail singular value distribution of CO-SVD as shown in Fig. \ref{eigen_dist}, and its strong advantage in compacting task-specific ability as validated in Fig. \ref{fig2}, the residual weight in our method for quantizing corresponds to the 
components that have small singular values and are redundant to the target task. Thus, quantizing this part is expected to induce minimal information loss. 
In the experiments, our method achieves better quantized adaptation performance than the compared methods.

\subsection{Extension to Vision Language Models}
\label{VLMA}

We further apply our method to the adaptation of LLaVA-1.5 \cite{liu2023visual,liu2024improved}, a representative vision language model. It integrates a vision tower with an LLM via a simple projection layer, enabling vision instruction tuning that achieves strong performance on visual question answering. 
However, vision language models often face a decline in their zero-shot instruction-following capabilities when fine-tuned on new domains \cite{zhai2023investigating,he2023continual,zhu2024model}.
We apply our method in knowledge-preserved adaptation mode to the fine-tuning of LLaVA-1.5, by sampling images and questions from the VQA datasets that need preservation. In this case, the covariance matrix of each linear layer reflects activations induced by both visual and text tokens. 
We show that our KPM helps alleviate performance degradation on zero-shot benchmarks during fine-tuning on a new task, without sacrificing adaptation performance compared to LoRA and full fine-tuning.

\begin{table*}[t!]  
	\caption{The experimental results of CorDA and CorDA++ in knowledge-preserved mode (KPM). We fine-tune LLaMA-2-7B on (a) Math, (b) Code, and (c) Instruction Following tasks. The rank $r$ of LoRA and CorDA is 128, and CorDA++ dynamically allocates rank based on its equivalent parameter budget. 
	CorDA and CorDA++ collect covariance matrices from NQ open. 
	All methods are implemented under the same training and evaluation settings. 
	The row of ``Zero-shot'' shows the pre-trained model's performance on the knowledge benchmarks. ``Avg.'' denotes the average score to reflect overall performance. 
	} 
\begin{center}
	\vspace{-2mm}
	\subcaption{Math}
	\vspace{-2mm}
		\begin{tabular}{l|c|ccc|cc|c}
			\toprule[1.5pt]
			Method & \#Params & Trivia QA & NQ open & WebQS & GSM8k & Math & Avg. \\
			\midrule
			Zero-shot & - & 52.51	& 14.99	& 5.86 & - & - & - \\
			\midrule
			Full fine-tuning & 6738M 					& 43.64$_{\pm\text{0.68}}$									& 3.13$_{\pm\text{1.02}}$							& 6.35$_{\pm\text{0.13}}$							  & \textbf{48.90}$_{\pm\text{0.49}}$			& \textbf{7.48}$_{\pm\text{0.22}}$ & 21.90 \\
			LoRA \cite{hu2022lora} & 320M		& 44.17$_{\pm\text{0.40}}$			 						  & 1.91$_{\pm\text{0.74}}$						& 6.64$_{\pm\text{0.13}}$									& 42.68$_{\pm\text{0.54}}$							& 5.92$_{\pm\text{0.15}}$ & 20.26\\
			CorDA & 320M 	  								& 	{44.30}$_{\pm\text{0.22}}$ 								& {9.36}$_{\pm\text{0.16}}$					& \textbf{7.14}$_{\pm\text{0.26}}$	  					& 44.58$_{\pm\text{0.33}}$	  						& 6.92$_{\pm\text{0.13}}$ 							 & {22.46}\\
			CorDA++ & 320M 								&	 \textbf{46.37}$_{\pm\text{0.31}}$					  &\textbf{12.02}$_{\pm\text{0.34}}$		&7.13$_{\pm\text{0.20}}$								&45.13$_{\pm\text{0.39}}$							& 7.05$_{\pm\text{0.11}}$   &   \textbf{23.54} \\
			\bottomrule[1.5pt]
		\end{tabular}\label{KPA_result-math}
	\vspace{2mm}
	
	\subcaption{Code}
	\vspace{-2mm}
		\begin{tabular}{l|c|ccc|cc|c}
			\toprule[1.5pt]
			Method & \#Params & Trivia QA & NQ open & WebQS & HumanEval & MBPP & Avg. \\
			\midrule
			Zero-shot & - & 52.51	& 14.99	& 5.86 & - & - & - \\
			\midrule
			Full fine-tuning & 6738M 				& 29.57$_{\pm\text{0.53}}$			 & 8.54$_{\pm\text{0.29}}$								& 4.76$_{\pm\text{0.37}}$						& \textbf{20.52}$_{\pm\text{0.29}}$	& \textbf{23.64}$_{\pm\text{0.38}}$    & 17.41 \\
			LoRA \cite{hu2022lora} & 320M 	& {51.42}$_{\pm\text{0.61}}$ 			& 9.30$_{\pm\text{0.21}}$							& 8.46$_{\pm\text{0.23}}$						& 16.8$_{\pm\text{0.38}}$					& 21.51$_{\pm\text{0.43}}$ 					& 21.50 \\
			CorDA & 320M 								& 50.02$_{\pm\text{0.33}}$ 				&	{11.72}$_{\pm\text{0.35}}$ 					& \textbf{8.56}$_{\pm\text{0.23}}$			& 18.36$_{\pm\text{0.19}}$					& 20.91$_{\pm\text{0.36}}$ 				   & {21.91}\\
			CorDA++ & 320M 							& \textbf{51.46}$_{\pm\text{0.25}}$ & \textbf{12.57}$_{\pm\text{0.37}}$			& 8.40$_{\pm\text{0.19}}$							& 19.23$_{\pm\text{0.40}}$				& 21.60$_{\pm\text{0.27}}$ 				      & \textbf{22.65} \\
			\bottomrule[1.5pt]
		\end{tabular}\label{KPA_result-code}
	
	\vspace{2mm}
	\subcaption{Instruction Following}
	\vspace{-2mm}
		\begin{tabular}{l|c|ccc|c|c}
			\toprule[1.5pt]
			Method & \#Params &Trivia QA & NQ open & WebQS & MTBench & Avg. \\
			\midrule
			Zero-shot & - & 52.51	& 14.99	& 5.86 & -  & - \\
			\midrule
			Full fine-tuning & 6738M 					& 26.6$_{\pm\text{1.39}}$ 				   & 8.45$_{\pm\text{0.38}}$						& 6.84$_{\pm\text{0.44}}$	& 4.85$_{\pm\text{0.09}}$ & 11.69 \\
			LoRA \cite{hu2022lora} & 320M 		& 47.46$_{\pm\text{0.52}}$				  & 10.28$_{\pm\text{0.22}}$						& 7.73$_{\pm\text{0.32}}$	& 4.60$_{\pm\text{0.14}}$ & 17.52 \\
			CorDA & 320M 									& {50.34}$_{\pm\text{0.58}}$ 			&	\textbf{14.43}$_{\pm\text{0.28}}$	& {8.17}$_{\pm\text{0.16}}$	& {5.05}$_{\pm\text{0.07}}$ & {19.50} \\
			CorDA++ & 320M 								& \textbf{51.36}$_{\pm\text{0.46}}$    & 14.37$_{\pm\text{0.19}}$						& \textbf{8.39}$_{\pm\text{0.21}}$& \textbf{5.13}$_{\pm\text{0.15}}$	& \textbf{19.81} \\
			\bottomrule[1.5pt]
		\end{tabular}\label{KPA_result-inst}
	
\end{center}
\label{KPA_result}
\vspace{-1mm}
\end{table*}

\section{Experiments}
\label{Experiments}

\subsection{Experimental Setup}

\textbf{Models, datasets, and benchmarks.}
In experiments, we fine-tune the pre-trained models including LLaMA-2-7B \cite{touvron2023llama}, LLaMA-2-13B,  LLaMA-3-8B, and Gemma-2-9B \cite{team2024gemma}, on Math, Code, and Instruction Following tasks. 
Following the settings in \cite{meng2024pissa}, we use MetaMathQA \cite{yu2024metamath} for fine-tuning on Math, and evaluate the model on GSM8k \cite{cobbe2021training} and Math \cite{yu2024metamath} validation sets. 
For Code, we train with the CodeFeedback \cite{zheng2024opencodeinterpreter} dataset, and test on the HumanEval \cite{chen2021evaluating} and MBPP \cite{austin2021program} benchmarks. 
Instruction following is trained on WizardLM-Evol-Instruct \cite{xu2024wizardlm} and evaluated on MTBench \cite{zheng2023judging}. 
For knowledge-preserved adaptation, the evaluation of pre-trained world knowledge is conducted by exact match scores (\%) on TriviaQA \cite{joshi-etal-2017-triviaqa}, NQ open \cite{lee-etal-2019-latent}, and WebQS \cite{berant2013semantic}.
For vision language model experiments, we fine-tune LLaVA-1.5 (Vicuna-7B) \cite{liu2023visual,liu2024improved} on OKVQA \cite{marino2019ok} and PMC-VQA \cite{zhang2023pmc}, and evaluate the decline of its zero-shot VQA abilities on benchmarks including VQAv2 \cite{goyal2017making}, VizWiz \cite{gurari2018vizwiz}, GQA \cite{hudson2019gqa}, SQA \cite{lu2022learn}, TextVQA \cite{singh2019towards}, MM-Bench \cite{liu2024mmbench}, and MM-Bench-CN \cite{zhang2023internlm}. 
For instruction-previewed adaptation, we also validate our method on nature language understanding, by {fine-tuning RoBERTa$_{\rm base}$ \cite{liu2019roberta} on the} General Language Understanding Evaluation (GLUE) \cite{wang2018glue} benchmark.

\textbf{Compared baselines.}
We denote our method in the original conference version as \emph{CorDA}, and refer to the improved version armed with the proposed dynamic covariance selection and dynamic rank allocation strategies as \emph{CorDA++}.  
Existing adapter initialization studies are mainly focused on improving fine-tuning performance, while pre-trained knowledge preservation has not been extensively studied. 
A higher downstream performance of one method is usually accompanied by a more severe zero-shot capability reduction. 
Therefore, for the experiments of knowledge-preserved adaptation, we only compare our method with full fine-tuning and LoRA. 
For instruction-previewed adaptation, we compare our method with strong baseline methods that improve LoRA from various perspectives, including adaptive rank (AdaLoRA \cite{zhang2023adaptive} and DyLoRA \cite{valipour2022dylora}), training strategies (LoRA+ \cite{lora+} and LoRA-FA \cite{zhang2023lora}), adapter structure (DoRA \cite{liu2024dora}), and initialization techniques (MiLoRA \cite{wang2024milora}, LoRA-GA \cite{wang2024lora}, PiSSA \cite{meng2024pissa}, and KaSA \cite{wang2025kasa}). 
For quantized adaptation experiments, we compare our quantized version, \emph{QCorDA}, with QLoRA \cite{dettmers2024qlora}, LoftQ \cite{li2024loftq}, and QPiSSA \cite{meng2024pissa}.

\textbf{Implementation details.} 
Following \cite{meng2024pissa}, we adopt the \verb*|bfloat16| data type for full fine-tuning, and use full precision for low rank adaptation experiments of both baseline and our methods. 
In quantized adaptation experiments, the residual weight is quantized to Normal Float 4-bit \cite{dettmers2022bit} with adapter in full precision. 
All methods are conducted under the same setup for fair comparison.  
For experiments on the GLUE benchmark, we adopt the setup in \cite{hu2022lora,gao2024parameterefficient,wang2025kasa}, which only fine-tunes \verb*|query| and \verb*|value| layers in each Transformer block. 
For the other experiments, we apply low-rank adapters to all linear layers.
When performing CorDA++, we set the default number of data sampling rounds, $N$, as 5 for dynamic covariance selection. We transform a specified LoRA rank into its corresponding learnable parameter budget, and apply Algorithm \ref{alg2} to get rank allocation across layers. 
We report the average result of five runs with different seeds. 
The complete training and evaluation details, baseline method introduction, and the reported metrics are provided in the Appendix.

\begin{figure*}[t]
	\centering
	\begin{subfigure}{0.495\textwidth}
		\centering
		\includegraphics[width=1.02\linewidth]{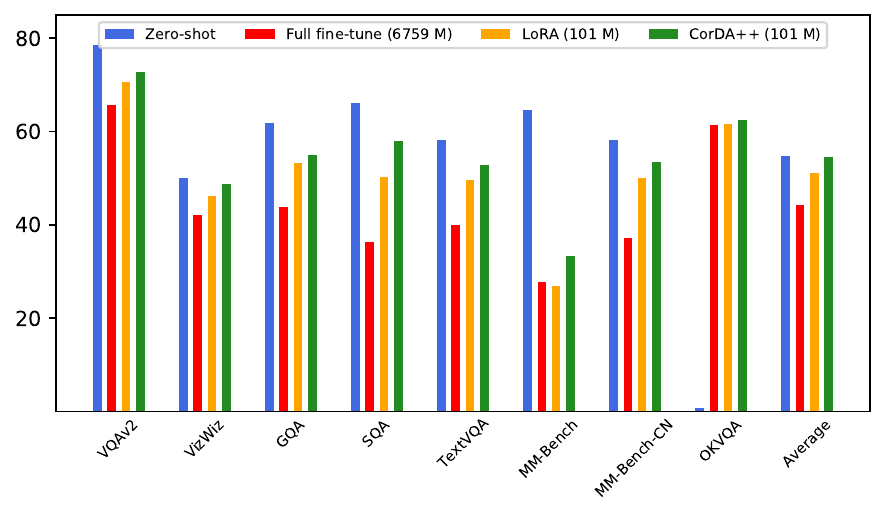}  
		\caption{Fine-tune on OKVQA}
		\label{okvqa}
	\end{subfigure}
	\begin{subfigure}{0.495\textwidth}
		\centering
		\includegraphics[width=1.02\linewidth]{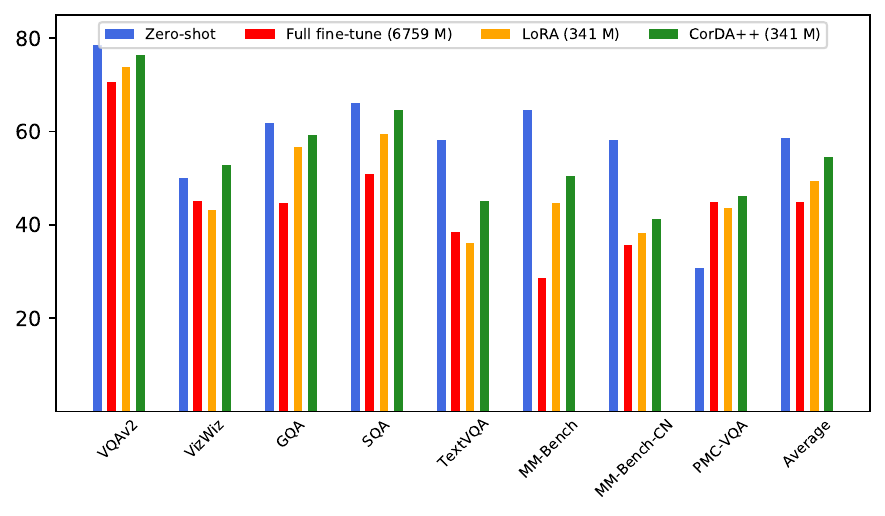}  
		\caption{Fine-tune on PMC-VQA}
		\label{pmcvqa}
	\end{subfigure}	
	\caption{Zero-shot and adaptation performances of LLaVA-1.5 using full fine-tuning, LoRA, and CorDA++ (KPM) on (a) OKVQA and (b) PMC-VQA. For LoRA and CorDA++, we use a rank of 32 for OKVQA and 128 for PMC-VQA.}
	\label{mllm}
\end{figure*}

\subsection{Results of Knowledge-Preserved Adaptation}

For our method in the knowledge-preserved adaptation mode (KPM), we collect covariance matrices by sampling 256 questions from the NQ open training set, which are then used to initialize adapters by Eq. (\ref{KPA_eq}).
As shown in Table \ref{KPA_result}, we fine-tune LLaMA-2-7B using full fine-tuning, LoRA, and our proposed CorDA and CorDA++ on Math, Code, and Instruction Following tasks. 
To assess both adaptability to new tasks and the retention of zero-shot capabilities, we measure fine-tuning performance alongside the scores on world knowledge benchmarks, and report their average values.
It is shown that full fine-tuning and LoRA both incur severe degradation on TriviaQA and NQ open after fine-tuned on the three tasks.
In particular, when fine-tuning on Math, there is almost a complete forgetting on NQ open (from 14.99 to 3.13 for full fine-tuning and to 1.91 for LoRA). 
A similar catastrophic forgetting can be also observed on TriviaQA for full fine-tuning in the Code and Instruction Following experiments. 
These results manifest the vulnerability of both LoRA and full fine-tuning to knowledge forgetting. 
In contrast, CorDA and CorDA++ achieve stronger fine-tuning performance than LoRA, while attaining the highest scores on all three world knowledge benchmarks. 
Notably, on the Instruction Following task, CorDA and CorDA++ even surpass full fine-tuning in the downstream performance measured by MTBench, with minimal compromise in the TriviaQA and NQ open capabilities. 
Compared with CorDA, CorDA++ yields a modest improvement in fine-tuning performance, while further mitigating the zero-shot ability reduction on TriviaQA and NQ open, which can be attributed to its improved compactness in the preserved components. 
The results of CorDA++ in KPM with other LLM scales and architectures are shown in Table~\ref{new_kpm_result}. 
Similar to LLaMA-2-7B, CorDA++ also consistently achieves the best average scores on LLaMA-2-13B, LLaMA-3-8B, and Gemma-2-9B.
These results reveal that CorDA++ in KPM can effectively alleviate knowledge forgetting during fine-tuning and improve the overall performance balancing adaptability and zero-shot capability maintenance.

\begin{table}[t!]  
	\caption{Adaptation results of CorDA++ (KPM) on Math while preserving NQ open with various pre-trained models.}
	\begin{center}
			\begin{tabular}{l|l|c|cc|c}
				\toprule[1.5pt]
				Model  & Method & NQ open & GSM8k & Math &  Avg.\\
				\midrule
				\multirow{3}{*}{LLaMA-2-13B} & Zero-shot  & 23.63 & - & - & - \\
				& LoRA~\cite{hu2022lora}	&	16.26	& 57.24	& 8.92	& 27.47 \\
				& CorDA++	& \textbf{20.05} & \textbf{60.18}	& \textbf{10.13}  & \textbf{30.12} \\
				\midrule
				\multirow{3}{*}{LLaMA-3-8B} & Zero-shot& 13.41 & - & - & - \\
				& LoRA~\cite{hu2022lora}	&	8.75	& 72.33	& 24.04	& 35.04 \\
				& CorDA++ &  \textbf{9.85}	& \textbf{75.01} & \textbf{25.50} & \textbf{36.79} \\
				\midrule
				\multirow{3}{*}{Gemma-2-9B} & Zero-shot& 12.85 & - & - & - \\
				& LoRA~\cite{hu2022lora}	&	9.28 &	83.47	&42.30	&45.02 \\
				& CorDA++ &  \textbf{10.49} & \textbf{84.42} & \textbf{42.61} & \textbf{45.84} \\		
				\bottomrule[1.5pt]
			\end{tabular}
		\label{new_kpm_result}
	\end{center}
\end{table}

We also apply CorDA++ to the fine-tuning of LLaVA-1.5 on OKVQA and PMC-VQA, which represent two different types of multimodal question answering tasks. 
OKVQA focuses on open-domain questions that require extensive commonsense and world knowledge, and PMC-VQA is a large scale medical VQA dataset. 
We fine-tune the whole model except the vision tower. When applying LoRA and CorDA++, we adopt a rank of 32 for OKVQA and 128 for PMC-VQA. 
We randomly sample 256 image-instruction pairs from all the evaluated zero-shot benchmarks to collect covariance matrices for our method. 
As shown in Fig. \ref{mllm}, full fine-tuning, LoRA, and CorDA++ in KPM achieve similar adaptation performance in the two tasks. 
However, both LoRA and full fine-tuning suffer from substantial degradation of zero-shot capabilities on the VQA benchmarks. 
As a comparison, CorDA++ better preserves zero-shot performance especially on the challenging benchmarks such as SQA and MM-Bench, which exhibit a large domain gap or broader task coverage compared to the fine-tuning task. 
It indicates that our method can also be applied to preserve multi-modal capabilities by jointly considering data context from visual and text domains.

\begin{figure*}[t]
	\centering
	\begin{subfigure}{0.496\textwidth}
		\centering
		\includegraphics[width=1\linewidth]{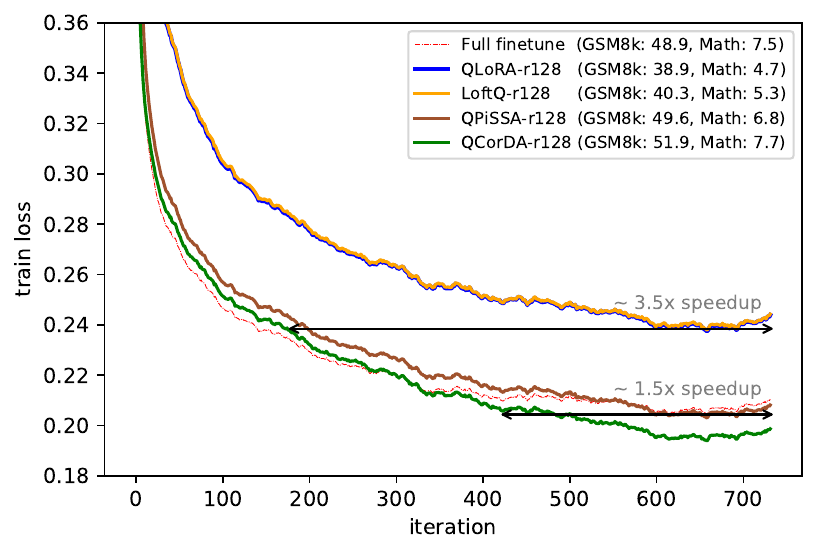}  
		\caption{$r=128$}
		\label{mathloss_r128}
	\end{subfigure}
	\begin{subfigure}{0.496\textwidth}
		\centering
		\includegraphics[width=1\linewidth]{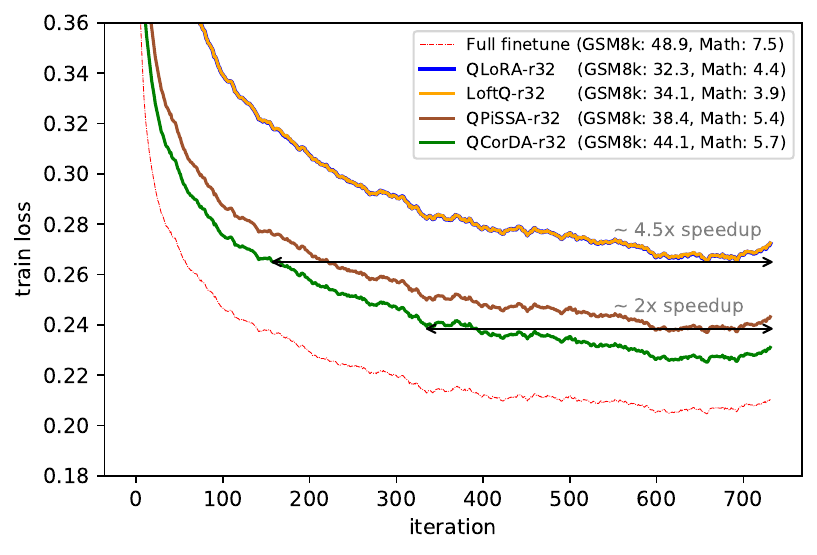}  
		\caption{$r=32$}
		\label{fmathloss_r32}
	\end{subfigure}	
	\caption{The training loss curves on MetaMath of full fine-tuning, QLoRA, LoftQ, QPiSSA, and QCorDA with (a) rank 128 and (b) rank 32. 
		The accuracies on GSM8k and Math are reported on the legends. 
		Smoothing is performed for the curves. 
		}
	\label{mathloss}
\end{figure*}

\begin{table*}[t!]  
	\renewcommand\arraystretch{1.0}
	\caption{The experimental results of CorDA and CorDA++ in the instruction-previewed adaptation mode (IPM) on Math, Code, and Instruction Following tasks using LLaMA-2-7B. We apply our methods by collecting covariance matrices from each of the fine-tuning datasets (MetaMathQA, CodeFeedback, and WizardLM-Evol-Instruct) for the three tasks, respectively. All methods are implemented under the same training and evaluation settings with LoRA adapter rank as 128.}
	\begin{center}
			\begin{tabular}{l|c|cc|cc|c|c}
				\toprule[1.5pt]
				Method  & \#Params & GSM8k & Math & HumanEval & MBPP & MTBench & Avg.\\
				\midrule
				Full fine-tuning & 6738M& 48.9$_{\pm\text{0.49}}$ 	& 7.48$_{\pm\text{0.22}}$ 	&{20.52}$_{\pm\text{0.29}}$ 	& {23.64}$_{\pm\text{0.38}}$ 	& 4.85$_{\pm\text{0.09}}$  & 21.08 \\ 
				LoRA \cite{hu2022lora} & 320M & 42.68$_{\pm\text{0.54}}$ 	& 5.92$_{\pm\text{0.15}}$ 	& 16.80$_{\pm\text{0.38}}$ & 21.51$_{\pm\text{0.43}}$ 	& 4.60$_{\pm\text{0.14}}$  & 18.30 \\
				AdaLoRA \cite{zhang2023adaptive} & 320M & 41.95$_{\pm\text{0.90}}$ 	& 6.24$_{\pm\text{0.38}}$ 	& 18.10$_{\pm\text{0.46}}$ 	& 20.19$_{\pm\text{0.71}}$ 	& 4.79$_{\pm\text{0.18}}$  & 18.25 \\
				DoRA \cite{liu2024dora} & 321M & 41.77$_{\pm\text{0.74}}$ 	& 6.20$_{\pm\text{0.48}}$  & 16.86$_{\pm\text{0.54}}$  & 21.60$_{\pm\text{0.49}}$  & 4.48$_{\pm\text{0.14}}$ & 18.18 \\
				MiLoRA \cite{wang2024milora} & 320M & 43.09$_{\pm\text{1.16}}$  & 6.31$_{\pm\text{0.39}}$  & 17.55$_{\pm\text{0.24}}$ 	& 20.22$_{\pm\text{0.37}}$ 	& 4.50$_{\pm\text{0.17}}$ & 18.33 \\
				LoRA+ \cite{lora+}& 320M & 47.84$_{\pm\text{0.39}}$  & 7.21$_{\pm\text{0.49}}$  & 20.07$_{\pm\text{0.38}}$  & 23.69$_{\pm\text{0.29}}$  & 5.11$_{\pm\text{0.06}}$ & 20.78\\
				LoRA-FA \cite{zhang2023lora} & 320M & 40.25$_{\pm\text{0.46}}$  & 5.66$_{\pm\text{0.47}}$  & 15.91$_{\pm\text{0.41}}$  & 20.01$_{\pm\text{0.32}}$  & 4.67$_{\pm\text{0.12}}$ & 17.30\\
				LoRA-GA \cite{wang2024lora} & 320M & 50.47$_{\pm\text{0.98}}$  & 7.13$_{\pm\text{0.44}}$ 	&	19.44$_{\pm\text{0.45}}$ & 	23.05$_{\pm\text{0.40}}$  & 5.04$_{\pm\text{0.10}}$ & 21.03 \\
				PiSSA \cite{meng2024pissa} & 320M & 51.48$_{\pm\text{0.34}}$ 	& 7.60$_{\pm\text{0.18}}$ 	& 19.48$_{\pm\text{0.45}}$ 	& 23.84$_{\pm\text{0.46}}$ & 4.92$_{\pm\text{0.07}}$ & 21.46 \\
				CorDA & 320M & {53.90}$_{\pm\text{0.56}}$ 	& {8.52}$_{\pm\text{0.27}}$ 	& 21.03$_{\pm\text{0.37}}$ 	& 24.15$_{\pm\text{0.44}}$ & {5.15}$_{\pm\text{0.09}}$  & {22.55} \\
				CorDA++ & 320M & \textbf{55.03}$_{\pm\text{0.52}}$ 	& \textbf{8.95}$_{\pm\text{0.37}}$ 	& \textbf{21.76}$_{\pm\text{0.39}}$ 	& \textbf{24.74}$_{\pm\text{0.47}}$ 	& \textbf{5.64}$_{\pm\text{0.12}}$ & \textbf{23.22} \\
				\bottomrule[1.5pt]
			\end{tabular}
		\label{IPA_result}
	\end{center}
\end{table*}

\subsection{Results of Instruction-Previewed Adaptation}

When users have no concern for pre-trained knowledge preservation and only pursue a higher fine-tuning performance, our method in the instruction-previewed adaptation mode (IPM) will be preferable. 
In this mode, we sample 256 items of instructions and responses from the fine-tuning dataset to collect covariance matrices. 
The resulting principal components of our decomposition extract the most responsive components to the fine-tuning task, and serve as the initialized adapters by Eq. (\ref{IPA_eq}) to accelerate learning of the new task. 
As shown in Table \ref{IPA_result}, CorDA and CorDA++ in IPM largely enhance the performance on downstream tasks of Math, Code, and Instruction Following, compared to their KPM results. 
CorDA++ achieves the best evaluation performance on the five benchmarks, outperforming PiSSA, a strong baseline that also adopts the principal components but is based on the plain SVD without considering data context. 
Compared to CorDA, CorDA++ consistently improves the adaptation performance on the three tasks. 
This suggests that the two proposed strategies, which induce stronger compactness in the task-specific principal components, are helpful to both KPM and IPM.
The results of CorDA++ in IPM with various LLM scales and architectures are shown in Table~\ref{new_ipm_result}. 
These results corroborate the benefit of data context for LoRA adapter initialization, implying that pre-capturing task characteristic in adapters can accelerate fine-tuning convergence (loss curves are provided in the quantized adaptation experiments) and lead to better performance. 
Moreover, our task-aware adaptation provides users with the flexibility to choose between KPM and IPM according to practical demands, which is a unique feature among the compared methods.

\begin{table}[t]  
	\caption{Adaptation results of CorDA++ (IPM) on Math with various pre-trained models.}
	\begin{center}
			\begin{tabular}{l|l|cc}
				\toprule[1.5pt]
				Model  & Method  & GSM8k & Math \\
				\midrule
				\multirow{3}{*}{LLaMA-2-13B} & LoRA~\cite{hu2022lora}  & 57.24 &  8.92 \\
				& PiSSA~\cite{meng2024pissa}	&	60.88	& 11.08 \\   
				& CorDA++ 	& \textbf{62.73} & \textbf{11.57} \\
				\midrule
				\multirow{3}{*}{LLaMA-3-8B} & LoRA~\cite{hu2022lora} & 72.33 &	24.04 \\
				& PiSSA~\cite{meng2024pissa}  & 76.80 &	26.26  \\
				& CorDA++ &  \textbf{77.34}	& \textbf{26.88}	 \\			
				\midrule			
				\multirow{3}{*}{Gemma-2-9B} & LoRA~\cite{hu2022lora} & 83.47 &	42.30 \\
				& PiSSA~\cite{meng2024pissa}	&	84.23	& 43.52  \\
				& CorDA++ 	& \textbf{84.97}	& \textbf{43.91} \\
				\bottomrule[1.5pt]
			\end{tabular}
		\label{new_ipm_result}
	\end{center}
\end{table}

\begin{table}[t]  
	\caption{
		The experimental results of different methods on the GLUE benchmark using RoBERTa$_{\rm base}$. 
		We apply CorDA++ (IPM) by sampling from each of the fine-tuning sub-tasks.  
		Baseline results are from \cite{gao2024parameterefficient,wang2025kasa}.
		Low-rank adapter based methods adopt a rank of $8$.
		Matthew's correlation and Pearson's correlation are the metrics of CoLA and STS-B, respectively. 
		The metric of the other tasks is accuracy.
		}
	\begin{center}
		\resizebox{\linewidth}{!}{
			\begin{tabular}{l|c|cccccc|c}
				\toprule[1.5pt]
				Method  & \#Params & SST-2 & MRPC & CoLA & QNLI & RTE & STS-B & Avg.\\
				\midrule
				Full fine-tuning & 125M & 94.8 & 90.2 & 63.6 &  92.8	& 78.7	&  91.2	& 85.2 \\
				BitFit \cite{bitfit} & 0.1M & 93.7 & \textbf{92.7} & 62.0 & 91.8 & 81.5 & 90.8 & 85.4\\
				Adapter$^D$ \cite{adapterdrop} &0.3M & 94.2 & 88.5 & 60.8 & 93.1 & 71.5 & 89.7 & 83.0\\
				Adapter$^D$ \cite{adapterdrop} &0.9M & 94.7 & 88.4 & 62.6 & 93.0 & 75.9 & 90.3 & 84.2\\
				LoRA \cite{hu2022lora} & 0.3M & 95.1 & 89.7 & 63.4 & \textbf{93.3} & 78.4 & 91.5 & 85.2\\
				AdaLoRA \cite{zhang2023adaptive}& 0.3M & 94.5 & 88.7 & 62.0 & 93.1 & 81.0 & 90.5 & 85.0\\
				DyLoRA \cite{valipour2022dylora} & 0.3M & 94.3 & 89.5 & 61.1 & 92.2 & 78.7 & 91.1 & 84.5\\
				DoRA \cite{liu2024dora} & 0.31M & 94.7	& 90.0	& 63.2 & 92.8 & 77.8& 91.2 & 85.0 \\
				PiSSA \cite{meng2024pissa} & 0.3M & 95.0 & 88.2 & 65.5 & 92.0 & 75.1 & 90.4 & 84.4\\
				MiLoRA \cite{wang2024milora} & 0.3M & 94.6 & 88.7 & 63.1 & 92.8 & 80.5 & 91.3 & 85.2\\
				KaSA \cite{wang2025kasa} & 0.3M & \textbf{95.2} & 90.7 & 65.8 & \textbf{93.3} & 81.6 & 91.1 & 86.3\\
				CorDA++ & 0.3M & 95.0	& {91.6} & \textbf{66.2} & 92.7 &  \textbf{81.9}	& \textbf{91.6} & \textbf{86.5}\\
				\bottomrule[1.5pt]
			\end{tabular}
		}
		\label{GLUE}
	\end{center}
	\vspace{-2mm}
\end{table}

In addition to the generation and instruction following tasks, we also apply our method to the General Language Understanding Evaluation (GLUE) benchmark by fine-tuning the RoBERTa$_{\rm base}$ model. 
We sample training data from each evaluated sub-task to apply our CorDA++ in IPM. 
Following \cite{hu2022lora,gao2024parameterefficient,wang2025kasa}, we fine-tune \verb*|query| and \verb*|value| layers and adopt a LoRA rank of 8. 
As shown in Table \ref{GLUE}, CorDA++ achieves the best performance on the CoLA, RTE, and STS-B datasets, and yields the highest average performance. 
These results demonstrate the scalability of our method to a broader range of LLM tasks.

\begin{table}[t!]
	\caption{Experimental results of QLoRA, LoftQ, QPiSSA, and QCorDA on quantized adaptation with various pre-trained models on Math and Code. All methods quantize the residual weight into Normal Float 4-bit (NF4) with adapters in full precision, and are trained and evaluated under the same setup.}
	\begin{center}
		\resizebox{\linewidth}{!}{
			\begin{tabular}{l|l|c|cc|cc}
				\toprule[1.5pt]
				Model & Method & rank & GSM8k & Math & HumanEval & MBPP \\
				\midrule
				\multirow{4}{*}{LLaMA-2-7B} & QLoRA \cite{dettmers2024qlora} & 128 & 38.89 & 4.7 & 15.59 & 19.29 \\
				& LoftQ \cite{li2024loftq} & 128  & 40.26 & 5.3 & 16.37 & 20.16 \\
				& QPiSSA \cite{meng2024pissa} & 128 & 49.58 & 6.8 & \textbf{21.34} & \textbf{24.56} \\
				& QCorDA  & 128 & \textbf{51.86} & \textbf{7.68} & {20.54} & 24.0 \\
				\midrule
				\multirow{4}{*}{LLaMA-3-8B} & QLoRA \cite{dettmers2024qlora} & 128 & 70.51 & 22.04 & 39.84 & 42.92 \\
				& LoftQ \cite{li2024loftq} & 128  & 68.23 & 22.0 & 41.85 & \textbf{44.57} \\
				& QPiSSA \cite{meng2024pissa} & 128 & 75.51 & 24.38 &42.44 & {43.11} \\
				& QCorDA  & 128 & \textbf{77.86} & \textbf{26.78} & \textbf{46.64} & 41.77 \\
				\midrule
				\multirow{12}{*}{Gemma-2-9B} & QLoRA \cite{dettmers2024qlora} & 128 & 80.82& 38.94 & 42.46 & 46.47 \\
				& LoftQ \cite{li2024loftq} & 128  & 79.98 & 38.16 & 43.05 & 45.73 \\
				& QPiSSA \cite{meng2024pissa} & 128 & 82.94 &\textbf{39.98} & 45.53 & 46.65 \\
				& QCorDA  & 128 & \textbf{83.17} & 38.74 & \textbf{53.4} & \textbf{47.12} \\
				\cmidrule(r){2-7}
				& QLoRA \cite{dettmers2024qlora} & 64 & 79.91 & 38.72 & 40.82 & 45.6 \\
				& LoftQ \cite{li2024loftq} & 64  & 79.76 & 37.42 & 40.98 & 44.63 \\
				& QPiSSA \cite{meng2024pissa} & 64 & 80.36 & 39.42 & 43.81 & 46.35 \\
				& QCorDA  & 64 & \textbf{81.8} & \textbf{40.02} & \textbf{43.94} & \textbf{46.95} \\
				\cmidrule(r){2-7}
				& QLoRA \cite{dettmers2024qlora} & 32 & 79.05 & 38.02 & 39.9 & 45.52 \\
				& LoftQ \cite{li2024loftq} & 32  & 78.92 & 37.6 & 33.55 & 44.68 \\
				& QPiSSA \cite{meng2024pissa} & 32 & 80.82&39.28 & 41.33 & \textbf{46.99} \\
				& QCorDA  & 32 & \textbf{81.96} & \textbf{40.24} & \textbf{42.52}  & 46.8\\
				\midrule
				\multirow{4}{*}{LLaMA-2-13B} & QLoRA \cite{dettmers2024qlora} &128& 52.16& 7.72 & 19.62 & 23.51 \\
				& LoftQ \cite{li2024loftq} & 128  & 51.39 & 7.18 & 16.13 & 14.57 \\
				& QPiSSA \cite{meng2024pissa} & 128 & 60.35 & 10.22 & 23.45 & 28.55 \\
				& QCorDA  & 128 & \textbf{62.4} & \textbf{11.92} & \textbf{27.15}  & \textbf{30.75}\\
				\bottomrule[1.5pt]
			\end{tabular}
		}
	\end{center}
	\label{quantized_adaptation}
\end{table}

We further combine our method in IPM with quantization as introduced in Sec. \ref{QA}. 
Concretely, we compare our quantized version, QCorDA, with representative methods including QLoRA, LoftQ, and QPiSSA, by fine-tuning various LLM models and scales with different ranks on Math and Code. 
The four methods only differ in adapter initialization, and are trained and evaluated under the same setup for fair comparison. 
As shown in Table \ref{quantized_adaptation}, QCorDA consistently outperforms all compared baselines in 19 out of the 24 evaluation scenarios. 
The train loss curves of the four methods when fine-tuning LLaMA-2-7B on Math are depicted in Fig. \ref{mathloss}. 
It is shown that QCorDA converges faster and attains a lower training loss compared with the baseline methods. 
When $r=128$, QCorDA achieves about 3.5x speedup over QLoRA and LoftQ, and 1.5x over QPiSSA. 
The gap becomes more evident at $r=32$, showing a 4.5x speedup over QLoRA and LoftQ, and 2x over QPiSSA. 
This can be attributed to QCorDA's superiority in concentrating task-specific capabilities into its principal components. 
At lower ranks, the initialized adapters of QCorDA correspond to the directions in the decomposed subspace that are most responsive to the target task, which is in line with our CO-SVD analysis in Sec. \ref{dist_analysis}.

\subsection{Ablation Studies}
\label{discussion}

We conduct ablation studies to examine the impact of data context selection and the way to initialize adapters in Table \ref{ablation}.
The first row replicates the PiSSA approach, where the top $r$ singular vectors and values from the plain SVD are used to initialize adapters.
In the second row, we replace the top $r$ components with the bottom $r$ ones for adapter initialization and observe no considerable gains on world knowledge benchmarks, suggesting that the plain SVD lacks the ability to extract task-specific components. 
As a result, freezing these task-agnostic principal components does not necessarily contribute to knowledge preservation. 
When our context-oriented decomposition (CO-SVD) is adopted with Wikitext-2, which is not closely related to question answering, the performance on knowledge benchmarks is much improved. 
Furthermore, using covariance matrices from actual QA datasets, such as TriviaQA and NQ open, leads to significant performance improvements on the three knowledge datasets, along with better average scores. 
Since TriviaQA and NQ open are datasets of the same task type, they produce close knowledge preserving and fine-tuning performance. 
Similarly, as shown in Table \ref{ablation_mtbench}, CorDA in IPM also results in close performance on MTBench when using data context from WizardLM-Evol-Instruct and Alpaca, both of which belong to the instruction following task. 
These results indicate that data context exhibits generalizability and serves as a valuable tool to orientate weight decomposition, enabling the extraction of task-specific components such that we can better preserve their associated knowledge or adapt them to a new task.

\begin{table}[t!]  
	\renewcommand\arraystretch{1.1}
	\caption{Ablation study for data context choice and adapter building manner by fine-tuning LLaMA-2-7B on Math. 
		The first row corresponds to the result of PiSSA that performs plain SVD and uses the top $r$ components to initialize adapters.
		CO-SVD+ adopts dynamic rank allocation and CO-SVD++ is with both proposed strategies.
		}	
	\begin{center}
		\resizebox{\linewidth}{!}{
			\begin{tabular}{l|cc|ccc|cc|c}
				\toprule[1.5pt]
				Method & Context & Adapter & TriviaQA & NQ open & WebQS & GSM8k & Math & Avg. \\
				\midrule
				Plain SVD & none & largest $r$ & 39.71$_{\pm\text{0.26}}$	& 1.02$_{\pm\text{0.23}}$	&6.30$_{\pm\text{0.39}}$	&51.48$_{\pm\text{0.34}}$	&7.60$_{\pm\text{0.18}}$& 21.22 \\
				Plain SVD & none & smallest $r$ & 39.94$_{\pm\text{0.17}}$ & 4.21$_{\pm\text{0.41}}$ & 6.25$_{\pm\text{0.17}}$ & 43.29$_{\pm\text{0.37}}$  & 5.96$_{\pm\text{0.13}}$ & 19.93 \\
				CO-SVD & Wikitext-2 & smallest $r$ &42.93$_{\pm\text{0.13}}$	& 7.20$_{\pm\text{0.15}}$	& 6.40$_{\pm\text{0.27}}$ & 42.99$_{\pm\text{0.34}}$ & 5.80$_{\pm\text{0.09}}$ & 21.06 \\
				CO-SVD &TriviaQA& smallest $r$ & 44.59$_{\pm\text{0.34}}$	& 8.86$_{\pm\text{0.20}}$	& 7.53$_{\pm\text{0.14}}$	& 44.81$_{\pm\text{0.28}}$	& 6.84$_{\pm\text{0.16}}$ &22.53 \\
				CO-SVD & NQ open& smallest $r$ & 44.30$_{\pm\text{0.22}}$ & 9.36$_{\pm\text{0.16}}$	& 7.14$_{\pm\text{0.26}}$	& 44.58$_{\pm\text{0.33}}$	& 6.92$_{\pm\text{0.13}}$ & 22.46\\
				CO-SVD+ & NQ open& smallest $r$ & 45.08$_{\pm\text{0.42}}$ & 10.34$_{\pm\text{0.33}}$ 	& 7.02$_{\pm\text{0.19}}$ & 45.04$_{\pm\text{0.31}}$ 	& 6.88$_{\pm\text{0.27}}$  & 22.87\\
				CO-SVD++ & NQ open& smallest $r$ &  {46.37}$_{\pm\text{0.31}}$					  &{12.02}$_{\pm\text{0.34}}$		&7.13$_{\pm\text{0.20}}$								&45.13$_{\pm\text{0.39}}$							& 7.05$_{\pm\text{0.11}}$   &   {23.54} \\
				\bottomrule[1.5pt]
			\end{tabular}
		}
		\label{ablation}
	\end{center}
	\vspace{-2mm}
\end{table}

We also investigate the impact of the two proposed strategies, dynamic covariance selection and dynamic rank allocation, in both KPM and IPM. 
As shown in the last two rows of Table \ref{ablation}, applying dynamic rank allocation (CO-SVD+) improves both performance on the knowledge benchmarks and the downstream task, while CO-SVD++ with the two strategies primarily brings additional gains in knowledge preservation. 
This may be attributed to the fact that, in KPM, reasonable rank allocation simultaneously benefits the retention of pre-trained knowledge and adaptation to a new task, while dynamic covariance selection focuses only on selecting representative context from knowledge datasets. 
For IPM, both strategies contribute positively to improving the fine-tuning performance, as demonstrated in Table \ref{ablation_mtbench}. 

Additionally, we analyze the effect of the number of sampling rounds, \emph{i.e,} $N$ in Eq. (\ref{select_C}), and report the time consumption and GSM8k accuracy of QCorDA, QLoRA, and full fine-tuning in Table \ref{ablation_N}.
When $N=1$, which means dynamic covariance selection is disabled, the pre-processing time only constitutes a small portion of the total training time. 
But QCorDA still significantly outperforms full fine-tuning and QLoRA on GSM8k due to the CorDA design. 
Note that when implementing dynamic covariance selection, we need to perform an extra SVD to the covariance matrix itself to calculate the compactness metric. 
To reduce pre-processing cost, we exclude layers with high-dimensional covariance matrices. 
For example, in LLaMA-2-7B, we only apply SVD to $C$ in $4096\times4096$ and $11008\times4096$, but omit the one in $11008\times11008$.
It is shown that when dynamic covariance selection is enabled with $N=3$, the accuracy further improves from 50.44 to 51.35, suggesting the benefit of selecting from a few candidates.
Although increasing $N$ causes longer pre-processing time, the accuracy gains diminish at a large $N$.
It indicates that 
a small pool of covariance matrices
suffices to bring a substantial performance improvement, while keeping pre-processing overhead within an acceptable range.

\begin{table}[t!]  
	\renewcommand\arraystretch{1.1}
	\caption{
		Adaptation performance on Instruction Following with different data context and strategies. LLaMA-2-7B is fine-tuned in the instruction-previewed adaptation mode.
		CorDA+ only adopts dynamic rank allocation and CorDA++ is with both proposed strategies.
		}	
	\begin{center}
			\begin{tabular}{l|c|c}
				\toprule[1.5pt]
				Method & Context &  MTBench\\
				\midrule
				CorDA  & WizardLM-Evol-Instruct & 5.15  \\
				CorDA  & Alpaca & 5.06  \\
				CorDA+ & WizardLM-Evol-Instruct & 5.40  \\
				CorDA++ & WizardLM-Evol-Instruct  & 5.64 \\
				\bottomrule[1.5pt]
			\end{tabular}
		\label{ablation_mtbench}
	\end{center}
\end{table}

\begin{table}[t!]  
	\renewcommand\arraystretch{1.1}
	\caption{Pre-processing and training time (GPU hour) measured on an NVIDIA A100-SXM4 GPU and fine-tuning performance on GSM8k with LLaMA-2-7B.
	$N$ refers to the number of data sampling rounds required by the dynamic covariance selection strategy.
	}	
	\begin{center}
			\begin{tabular}{l|c|c|c}
				\toprule[1.5pt]
				Method & Pre-process time &  Train time & GSM8k\\
				\midrule
				Full fine-tune  & -& 17.3 & 48.90 \\ 
				QLoRA & - & 7.3 & 38.89\\
				\midrule
				QCorDA ($N$=1) & 0.15 & 7.3 & 50.44 \\
				QCorDA ($N$=3) & 0.64 & 7.3 & 51.35 \\
				QCorDA ($N$=5) & 1.06 & 7.3 & 51.86\\
				QCorDA ($N$=10) & 2.12 & 7.3 & 51.98\\
				\bottomrule[1.5pt]
			\end{tabular}
		\label{ablation_N}
	\end{center}
\end{table}

\section{Conclusion}

In this work, we present CorDA, a novel low-rank adaptation method that initializes LoRA adapters in a task-aware manner. 
By performing our proposed CO-SVD on the product of weight matrix and its corresponding covariance matrix, the embedded data context can orientate the decomposition, allowing the principal components to concentrate task-specific capability. 
Based on this design, we develop two complementary adaptation modes.
The knowledge-preserved mode mitigates forgetting of pre-trained knowledge by freezing the principal components, while the instruction-previewed mode adapts them to accelerate the learning of a new task. 
We further introduce a compactness metric, and propose two adaptive strategies, dynamic covariance selection and dynamic rank allocation based on the same metric. 
They provide each layer with a representative covariance matrix with strong contextual relevance, and a tailored rank allocation to improve parameter budget utilization. 
Experimental results demonstrate that our method in KPM outperforms LoRA not only in downstream performance but also in maintaining zero-shot capabilities 
for both large language models and vision language models. 
Meanwhile, the IPM exhibits superior fine-tuning performance and faster convergence in both standard and quantized adaptation across various tasks.

\bibliographystyle{IEEEtran}
\bibliography{corda}

\newpage
\
\newpage

 {\appendices
	
	\section{Experiment Details}

	\textbf{Baseline method introduction.}
	We introduce the baseline methods adopted in our experiments as follows:
	\begin{itemize}
		\item Full fine-tuning optimizes the whole pre-trained model during fine-tuning. 
		\item LoRA \cite{hu2022lora} introduces two low-rank adapter matrices for each linear layer and only optimizes the low-rank adapters during fine-tuning. 
		\item AdaLoRA \cite{zhang2023adaptive} adopts the SVD form for weight update and adaptively allocates the parameter budget through their proposed importance score. 
		\item  DoRA \cite{liu2024dora} normalizes the weight matrices and introduces an extra layer to learn magnitude, such that the LoRA adapters only learn directions. 
		\item MiLoRA \cite{wang2024milora} applies the plain SVD and adopts residual components as initialized adapters with principal components unchanged. 
		\item LoRA+ \cite{lora+} proposes to use different learning rates for the two low-rank adapters $A$ and $B$. 
		\item LoRA-FA \cite{zhang2023lora} proposes to only update the projection-up weight $B$ while freezing both pretrained weight $W$ and projection-down weight $A$.
		\item LoRA-GA \cite{wang2024lora} proposes to initialize low-rank adapters with the eigenvectors of the full gradient matrix.
		\item PiSSA \cite{meng2024pissa} applies the plain SVD and adopts the principal components as initialized adapters with residual weight frozen. 
		\item BitFit \cite{bitfit} only fine-tunes bias vectors with the remaining model parameters frozen. 
		\item Adapter$^D$ \cite{adapterdrop} proposes to drop adapters from lower Transformer layers during training and inference to improve efficiency. 
		\item DyLoRA \cite{valipour2022dylora} trains LoRA blocks for a range of ranks and proposes a search-free method to select the optimal rank.
		\item KaSA \cite{wang2025kasa} applies SVD truncation to the base weight and reparameterizes the updates in SVD form with regularizations of update direction and orthogonality.
		\item QLoRA \cite{dettmers2024qlora} quantizes the pre-trained weight into a lower bit, and keeps LoRA adapters in full precision. 
		\item LoftQ \cite{li2024loftq} finds a proper low-rank initialization by decomposing the quantization error matrix of QLoRA. 
		\item QPiSSA \cite{meng2024pissa} adopts the low-rank initialization of PiSSA and quantizes the residual weight. 
	\end{itemize}

	\textbf{Metrics.} 
	For knowledge benchmarks on TriviaQA, NQ open, WebQS, we report the exact match metric that measures the percentage of predictions matching any one of the ground truth answers exactly. 
	For evaluation on the Math task, we report the accuracy on the GSM8k and Math validation sets. 
	For the Code task, we report the ``Pass@1'' score that measures the percentage of test cases where the top-1 generated code passes unit tests. 
	For the Instruction Following task, we report the score on MTBench, which is the average score assigned by GPT-4 to a model's responses across 80 multi-turn dialogue tasks. 
	In the experiments of natural language understanding on the GLUE benchmark, the metrics of CoLA and STS-B are Matthew's correlation coefficient and Pearson's correlation coefficient, respectively. Accuracy is reported for the other sub-tasks of GLUE, and also all the visual question answering benchmarks. 
	In our experiments, we report the average result of five runs with different seeds.

	\textbf{Implementation details.} 
	We use the parameter-efficient fine-tuning framework developed by Hugging Face \footnote{\url{https://github.com/huggingface/peft}} to do the adaptation experiments. 
	For training on Math, Code, and Instruction Following, we adopt the same setup as PiSSA \cite{meng2024pissa}. 
	Concretely, full fine-tuning adopts the \verb*|bfloat16| data type, and LoRA-based methods use full precision for weights and adapters. 
	In quantized adaptation experiments, the residual weight is quantized to Normal Float 4-bit \cite{dettmers2022bit} with adapter in full precision.
	Optimization is performed with the AdamW optimizer, a batch size of 128, and a learning rate of $2e-5$. 
	We employ the cosine annealing schedule with a warmup ratio of 0.03 and do not apply weight decay. 
	Training is conducted exclusively on the first 100,000 conversations from the dataset for one epoch, with loss computation solely based on the response. 
	Our experiments are executed on the NVIDIA A100-SXM4(80GB) GPUs. 
	Publicly available platforms are utilized for the evaluation of world knowledge benchmarks (TriviaQA, NQ open, and Web QS) \footnote{\url{https://github.com/EleutherAI/lm-evaluation-harness}}, Code (HumanEval and MBPP) \footnote{\url{https://github.com/bigcode-project/bigcode-evaluation-harness}}, and Instruction Following (MTBench) \footnote{\url{https://github.com/lm-sys/FastChat}}. 
	Quantized adaptation experiments are conducted using the same training and evaluation recipe. 
	The number of steps for alternating optimization in LoftQ \cite{li2024loftq} is set as 5. 
	The default number of data sampling rounds for dynamic covariance selection is 5, with 256 samples in each round. 
	For our experiments of fine-tuning on Math, Code, and Instruction Following, we set LoRA $\alpha$ the same value as LoRA rank $r$, and apply low-rank adapters to all linear layers. 
	All compared methods are trained and evaluated under the same setup for fair comparison. 
	
	For vision language model experiments with LLaVA-1.5 (Vicuna 7B), we fine-tune the MLP connector and language model weights with vision tower parameters frozen for both full fine-tuning, LoRA, and our method. 
	We adopt the same optimizer, learning rate, batchsize, and warmup ratio as the language model experiments. 
	Training is performed with 100,000 question-answer pairs for one epoch on PMC-VQA, and 90,090 pairs on OKVQA. 
	
	For our experiments on the GLUE benchmark, following \cite{hu2022lora,wang2025kasa,gao2024parameterefficient}, we only fine-tune \verb*|query| and \verb*|value| layers in each Transformer block and adopt a LoRA rank $r$ of 8 and a LoRA $\alpha$ of 16. 
	We adopt the AdamW optimizer with a warmup ratio of 0.06, a linear learning rate schedule, and no weight decay. 
	The max sequence length is set as 512. 
	We train for 40 epochs for QNLI and STS-B, and 100 epochs for the other sub-tasks. 
	We use a batchsize of 128 for SST-2, and 32 for the other sub-tasks. 
	Learning rates are set as $3e-4$ for QNLI and STS-B, $4e-4$ for MRPC, CoLA, and RTE, and  $5e-4$ for SST-2. 
	To collect covariance matrices for our method, we perform sampling from the training set of each sub-task.

	\section{More Results}
	
	\begin{figure*}[t]
		\centering
		\begin{subfigure}{0.325\textwidth}
			\centering
			\includegraphics[width=1.\linewidth]{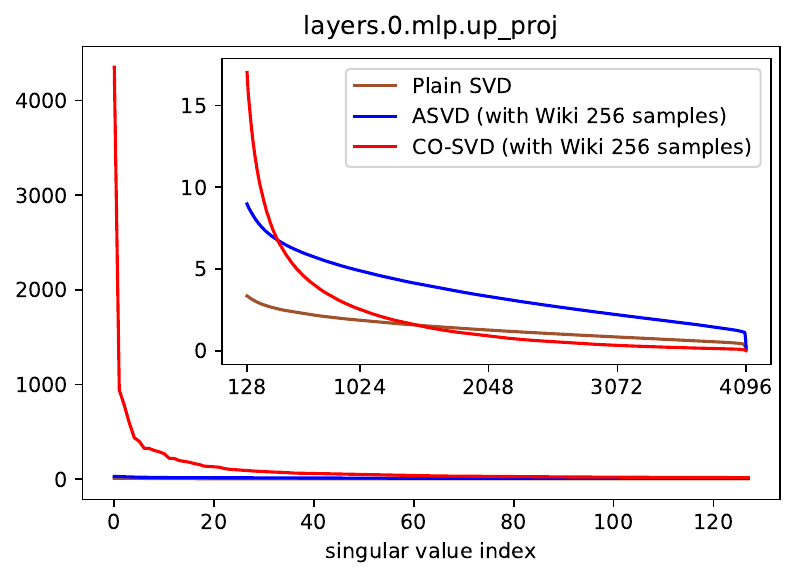}
			\label{eigen_5}
		\end{subfigure}
		\begin{subfigure}{0.325\textwidth}
			\centering
			\includegraphics[width=1.\linewidth]{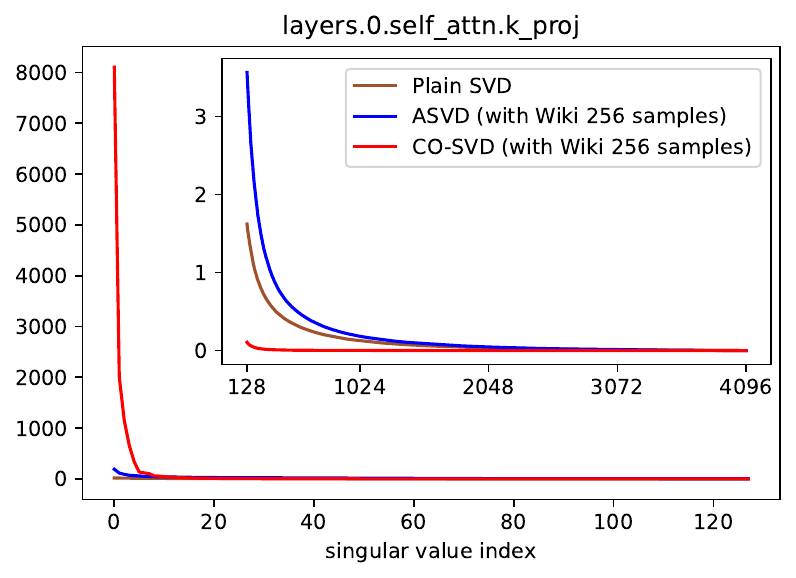}  
			\label{eigen_6}
		\end{subfigure}
		\begin{subfigure}{0.325\textwidth}
			\centering
			\includegraphics[width=1.\linewidth]{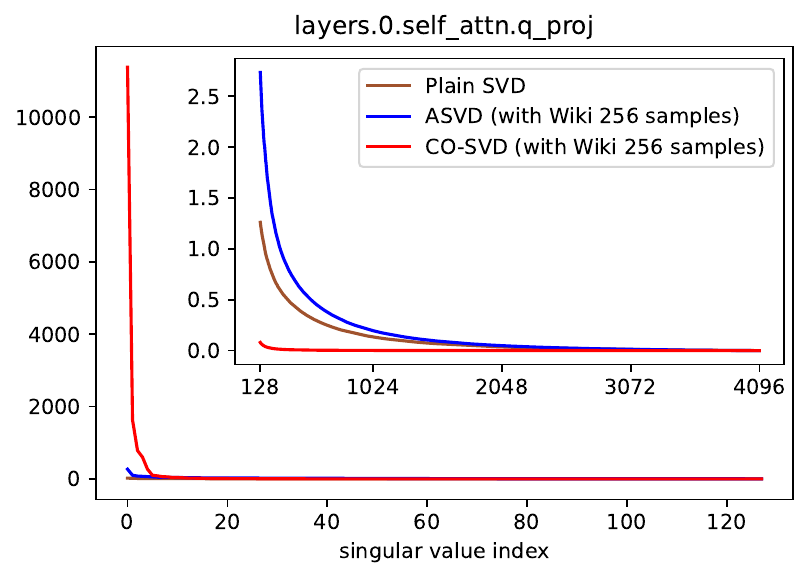}  
			\label{eigen_7}
		\end{subfigure}	
		\caption{Singular value distribution of plain SVD, ASVD, and our proposed context-oriented SVD (CO-SVD). The embedded figures show singular values with indices from 128 to 4096, and main figures illustrate the top 128 singular values. }
		\label{eigen_dist_appendix}
	\end{figure*}

	\begin{table}[t!]  
		\caption{Perplexity (lower is better) on Wikitext-2 and Penn TreeBank (PTB) after decomposing each weight matrix and discarding the bottom $r$ components using plain SVD, ASVD, and CO-SVD with various sample number and dataset choice.}
		\begin{center}
			\resizebox{\linewidth}{!}{
				\begin{tabular}{clcccccccc}
					\toprule[1.5pt]
					\multirow{2}{*}{Test Data} &\multirow{2}{*}{Method} & \multicolumn{8}{c}{The number of bottom components to discard ($r$).}\\
					&& 0 & 16 & 32 & 64 &128 & 256 & 512 & 1024 \\
					\midrule
					\multirow{5}{*}{Wikitext-2} & Plain SVD & 5.47	& 6.32	& 7.31	& 9.89	& 18.03	& 18.5	& 25.42	& 73.92 \\
					& ASVD \cite{yuan2023asvd} (with 256 Wiki samples)  & 5.47	& 6.08	& 6.67	& 7.86	& 8.71	& 9.92	& 12.37	& 20.34 \\
					& CO-SVD (with 32 Wiki samples) &5.47 &  5.48	& 5.48	& 5.49	& 5.52	& 5.58	& 5.79	& 6.62\\
					& CO-SVD (with 256 Wiki samples) &5.47 & 5.48	& 5.48	& 5.48	& 5.5	& 5.54	& 5.69	& \textbf{6.35}\\
					& CO-SVD (with 256 PTB samples) &5.47 & 5.49	& 5.5	& 5.52	& 5.57	& 5.74	& 6.25	& 8.69 \\
					\midrule
					\multirow{5}{*}{PTB}  & Plain SVD & 20.82 &	35.25	& 33.42	& 37.46	& 55.47 & 70.25	& 98.6	& 763.44\\
					& ASVD \cite{yuan2023asvd} (with 256 PTB samples)  & 20.84	& 33.42	& 32.05	& 31.67	& 35.36	& 40.23	& 51.28	& 93.42\\
					& CO-SVD (with 32 PTB samples) & 20.75 & 20.75	& 20.76	& 20.78	& 20.83	& 20.91	& 21.17	& 22.68\\
					& CO-SVD (with 256 PTB samples)	& 20.88 & 20.88	& 20.88	& 20.89	& 20.91	& 20.94	& 21.14	& \textbf{22.28}\\
					& CO-SVD (with 256 Wiki samples)	& 20.34 & 20.34	& 20.32	& 20.41 & 20.59	& 21.25	& 22.94	& 29.69\\
					\bottomrule[1.5pt]
				\end{tabular}
			}
			\label{fig2_appendix}
		\end{center}
	\end{table}

	The visualization results of singular value distribution of plain SVD, ASVD, and our proposed CO-SVD for more layers in the first block are shown in Fig. \ref{eigen_dist_appendix}. 
	The distributional characteristics are consistent with those of the other layers presented in the main paper.

	Table \ref{fig2_appendix} provides more results of Fig. 4 (in the main paper) by using different sample number and dataset choice. 
	It is shown that the number of sampled data only has a very limited impact. 
	When the smallest 1024 ranks are discarded, using 32 samples is slightly worse than 256 samples in both Wikitext-2 and PTB for our CO-SVD. 
	It implies that a small number of samples is enough to aggregate data context and compact task-specific capability into the principle components by CO-SVD. 
	Besides, collecting samples from the same dataset as the one used to test is able to attain a better performance especially at a large discarded rank. 
	For example, when discarding the bottom 1024 components, CO-SVD (with 256 Wiki samples) is better than CO-SVD (with 256 PTB samples) on Wikitext-2 (6.35 v.s. 8.69), and CO-SVD (with 256 PTB samples) is better than CO-SVD (with 256 Wiki samples) on PTB (22.28 v.s. 29.69). 
	This reveals that precisely capturing the data context in our decomposition helps to better associate the principal components with the target task capability, and also explains why our method is superior to other LoRA based methods that do not leverage data context.

	We provide more visualization results of the covariance matrices collected from three tasks MetaMath, NQ open, and Trivia QA in Fig. \ref{cov_visualization_2}.
	Since the original dimension in 4096 or 11008 will be too large to be informative, we downsample the covariance matrices into $32 \times 32$ and visualize their heatmaps. We provide the results from the activations before different linear layers including \verb|self_attn.k_proj| (the same as \verb|self_attn.q_proj| and \verb|self_attn.v_proj| due to the same input), \verb|self_attn.o_proj|, \verb|mlp.down_proj|, and \verb|mlp.gate_proj| (the same as \verb|mlp.up_proj|) in the first block (Fig. 1 in the main paper), and the \verb|self_attn.o_proj| layers in later blocks (Fig. \ref{cov_visualization_2}). 
	It is shown that the heatmaps from NQ open and TriviaQA (both are QA tasks) share similar patterns (marked in red circles), which do not appear in the heatmap from the different task MetaMath. 
	These visualization results empirically support the motivation that covariance matrix patterns can be used to characterize the task triggered by the input.
	We use such outlier distribution pattern to orientate the decomposition of pre-trained weights, leading to task-aware adapter initialization.

}

\begin{figure*}
	\centering
	\includegraphics[width=0.8\linewidth]{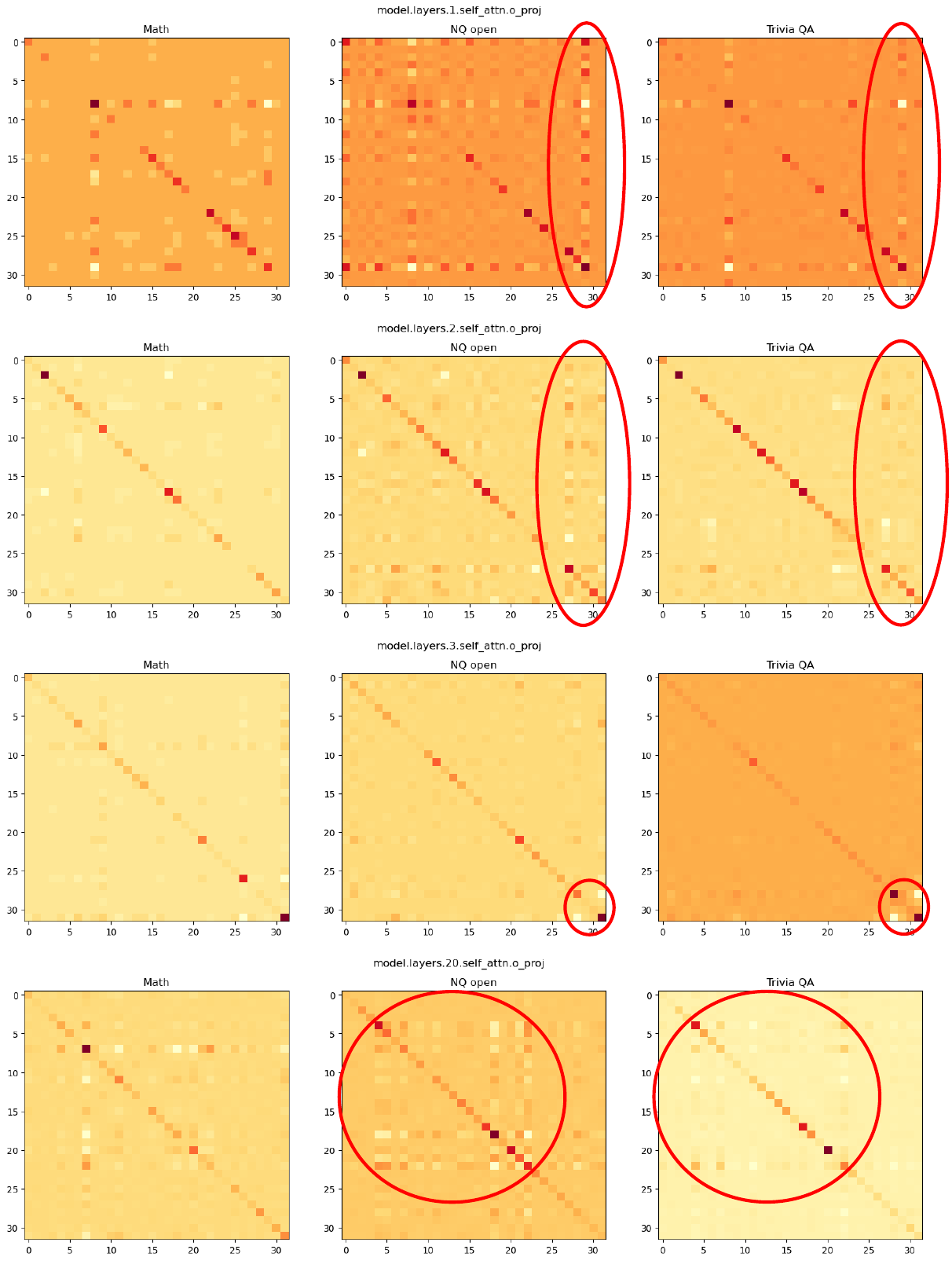}
	\caption{Covariance matrix visualization results of the ``self\_attn.o\_proj'' layers in blocks of different depths.}
	\label{cov_visualization_2}
\end{figure*}

\vfill

\end{document}